\documentclass[10pt,journal,compsoc]{IEEEtran}
\usepackage[pdftex]{graphicx}
\usepackage{booktabs, makecell, tabularx}
\usepackage{rotating}
\usepackage[export]{adjustbox}
\usepackage{gensymb}
\usepackage{ragged2e}
\usepackage{verbatim}
\usepackage{soul}
\usepackage{enumitem}
\ifCLASSOPTIONcompsoc
\usepackage[nocompress]{cite}
\else
\usepackage{cite}
\fi

\ifCLASSINFOpdf
\usepackage[pdftex]{graphicx}
\graphicspath{{./}}
\else
\fi
\usepackage[pagebackref=false,breaklinks=true,letterpaper=true,colorlinks,citecolor=blue,bookmarks=false]{hyperref}
\usepackage{epstopdf}
\usepackage{xspace}
\usepackage{pgfplots}
\usepackage{tikz}

\usepackage{amsmath}
\usepackage{amssymb}
\usepackage{array}
\ifCLASSOPTIONcompsoc
\usepackage[caption=false,font=footnotesize,labelfont=sf,textfont=sf]{subfig}
\else
\usepackage[caption=false,font=footnotesize]{subfig}
\fi

\ifCLASSOPTIONcaptionsoff
\usepackage[nomarkers]{endfloat}
\let\MYoriglatexcaption\caption
\renewcommand{\caption}[2][\relax]{\MYoriglatexcaption[#2]{#2}}
\fi
\usepackage{url}
\usepackage{etex}
\usepackage{diagbox}
\usepackage{multirow}
\usepackage{makecell}
\makeatletter
\def\@IEEEsectpunct{.\ \,}
\def\paragraph{\@startsection{paragraph}{4}{\z@}{1.5ex plus 1.5ex minus 0.5ex}%
	{0ex}{\normalfont\normalsize\sffamily\bfseries}}
\makeatother
\newcommand{\figref}[1]{Fig.~\ref{#1}}
\newcommand{\tabref}[1]{Tab.~\ref{#1}}

\newcommand{\equref}[1]{Eq. (\ref{#1})}

\def\ie{\emph{i.e.,~}}

\begin{document}
	
	\title{Unmixing Convolutional Features for \\ Crisp Edge Detection}
	
		\author{Linxi Huan,
		Nan Xue,
		Xianwei Zheng*,
		Wei He,
		Jianya Gong,
		Gui-Song Xia
		\IEEEcompsocitemizethanks{\IEEEcompsocthanksitem L. Huan and X. Zheng are with the LIESMARS, Wuhan University, Wuhan 430079, China. E-mail: \{whu\_hlx, zhengxw\}@whu.edu.cn.
			\IEEEcompsocthanksitem N. Xue and G.-S. Xia are with the School of Computer Science, Wuhan University, Wuhan 430079, China. 
			\IEEEcompsocthanksitem W. He is with the RIKEN Center, Tokyo 1030027, Japan. 
			\IEEEcompsocthanksitem J. Gong is with the LIESMARS and the School of Remote Sensing and Information Engineering, Wuhan University, Wuhan 430079, China.
			
			\IEEEcompsocthanksitem Linxi Huan and Nan Xue contributed equally to this work.
			
		    \IEEEcompsocthanksitem Corresponding author: X. Zheng.
		}
	}
	
	\markboth{Journal of \LaTeX\ Class Files,~Vol.XX, No.XX, 2020}%
	{Shell \MakeLowercase{\textit{et al.}}: Bare Advanced Demo of IEEEtran.cls for IEEE Computer Society Journals}
	
	\IEEEtitleabstractindextext{%
		\begin{abstract}		
	This paper presents a context-aware tracing strategy (CATS) for crisp edge detection with deep edge detectors, based on an observation that the localization ambiguity of deep edge detectors is mainly caused by the mixing phenomenon of convolutional neural networks: \emph{feature mixing} in edge classification and \emph{side mixing} during fusing side predictions.
	The CATS consists of two modules: a novel {\em tracing loss} that performs feature unmixing by tracing boundaries for better side edge learning, and a {\em context-aware fusion block} that tackles the side mixing by aggregating the complementary merits of learned side edges.
	Experiments demonstrate that the proposed CATS can be integrated into modern deep edge detectors to improve localization accuracy. With the vanilla VGG16 backbone, in terms of BSDS500 dataset, our CATS improves the F-measure (ODS) of the RCF and BDCN deep edge detectors by 12\% and 6\% respectively when evaluating without using the morphological non-maximal suppression scheme for edge detection.
\end{abstract}
		
		}

	\maketitle

	\IEEEdisplaynontitleabstractindextext
	
	\IEEEpeerreviewmaketitle

	\ifCLASSOPTIONcompsoc
	\IEEEraisesectionheading{\section{Introduction}\label{sec:introduction}}
	\else
	\section{Introduction}
	\label{sec:introduction}
	\fi
	
	\IEEEPARstart{A}{s} a fundamental computer vision task of localizing boundaries of perceptually salient objects in natural images, edge detection has been studied with a long history~\cite{roberts1963machine}. In the early stage, the low-level features were extensively studied for detecting edges~\cite{canny1986a, kittler1983accuracy, martin2004learning, xiaofeng2012discriminatively}. These approaches can obtain crisp edge maps but leave a challenging problem of suppressing high-frequency texture regions. Later, \textcolor{black}{learning techniques were introduced to classify image patches as edge and non-edge classes~\cite{Wei2015DeepContour, Bertasius2015DeepEdge}.} End-to-end deep edge detectors~\cite{xie2017hed, liu2019richer, he2019bi-directional} have been recently proposed to learn multi-level side edges with deep supervision, and weight the side predictions according to the learned side importance to obtain the final predictions.

\begin{figure*}[!t]
	\setlength{\fboxsep}{0pt}
	\begin{tabular}{cc|ccccc}
		\multicolumn{2}{c}{}& \multicolumn{5}{c}{Ground-Truth Edge Maps by Different Annotators}\\
		\rotatebox{90}{~~~~Image}&
		\includegraphics[width=0.13\linewidth]{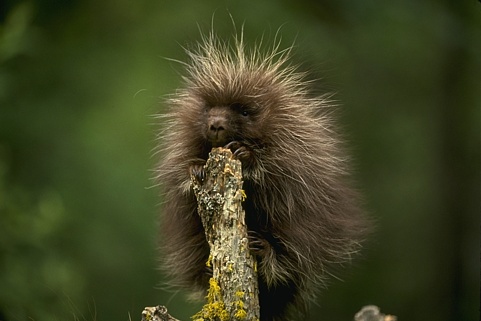}&
		\fbox{\includegraphics[width=0.13\linewidth]{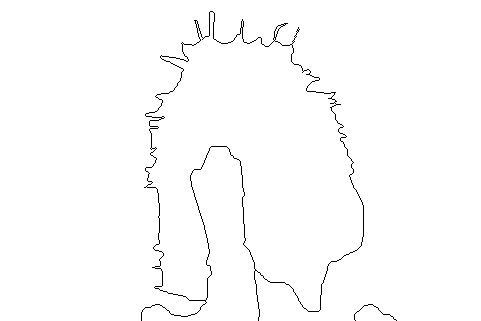}}&
		\fbox{\includegraphics[width=0.13\linewidth]{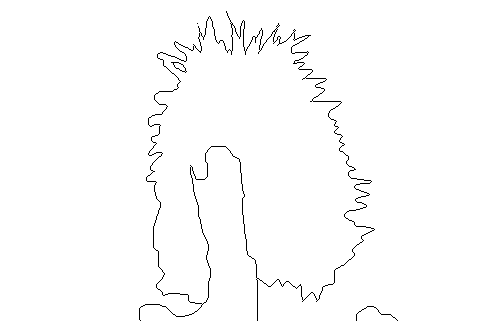}}&
		\fbox{\includegraphics[width=0.13\linewidth]{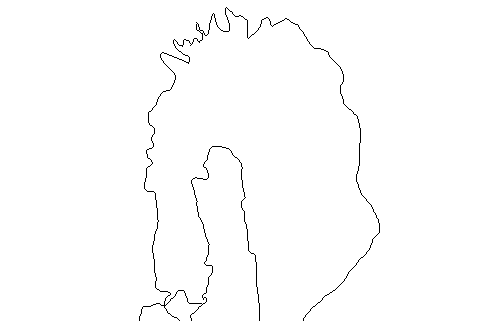}}&
		\fbox{\includegraphics[width=0.13\linewidth]{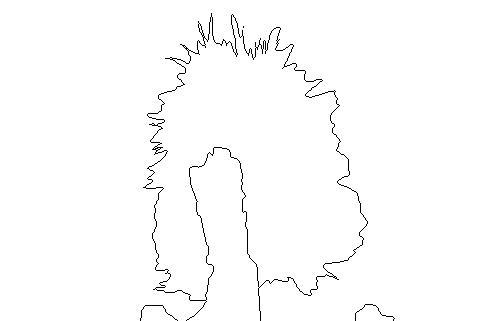}}&
		\fbox{\includegraphics[width=0.13\linewidth]{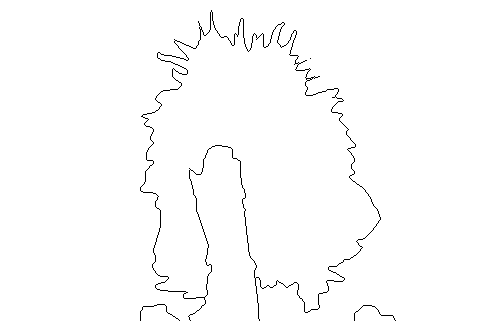}}
		\\
		& Final Edge & Side Edge$_1$ & Side Edge$_2$ & Side Edge$_3$ & Side Edge$_4$ & Side Edge$_5$\\ 
		\rotatebox{90}{~~~~~HED} &  \fbox{\includegraphics[width=0.13\linewidth]{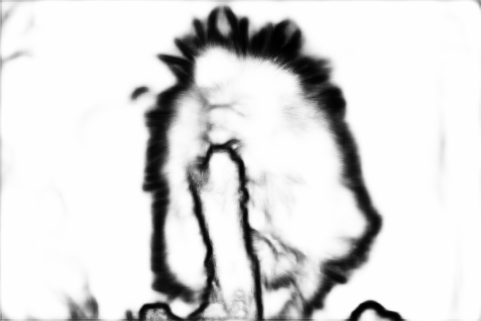}}
		& 
		\fbox{\includegraphics[width=0.13\linewidth]{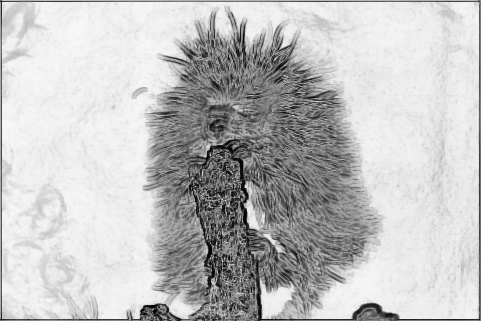}}
		&
		\fbox{\includegraphics[width=0.13\linewidth]{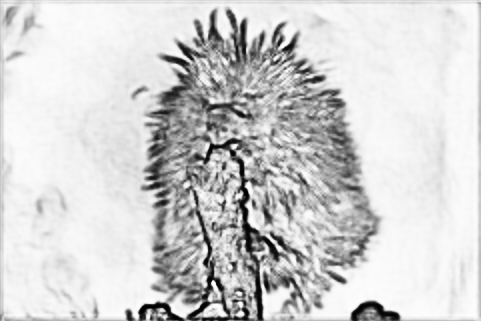}}
		&
		\fbox{\includegraphics[width=0.13\linewidth]{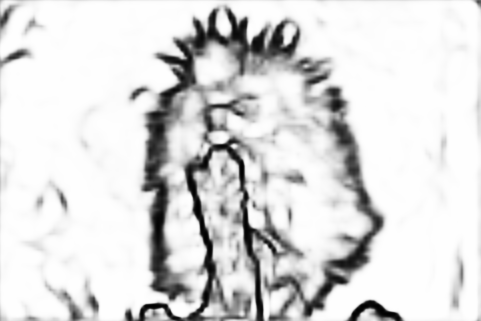}}
		&
		\fbox{\includegraphics[width=0.13\linewidth]{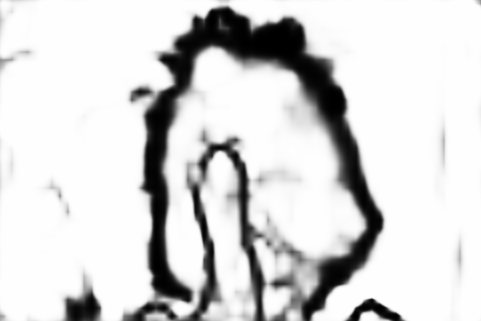}}
		&
		\fbox{\includegraphics[width=0.13\linewidth]{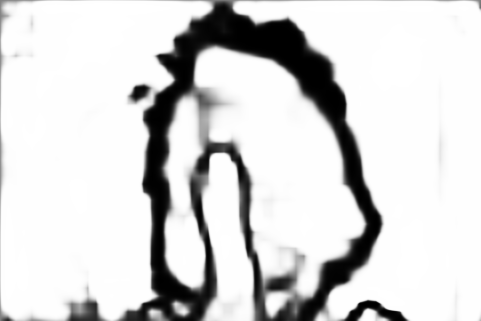}}\\
		\rotatebox{90}{~~~~~RCF} &  \fbox{\includegraphics[width=0.13\linewidth]{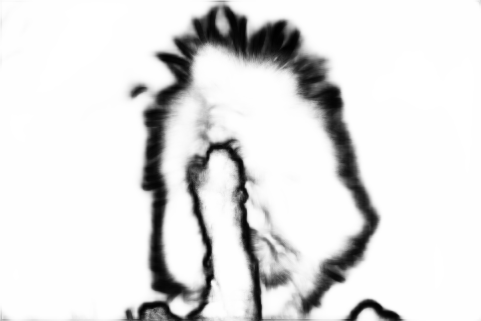}}
		& 
		\fbox{\includegraphics[width=0.13\linewidth]{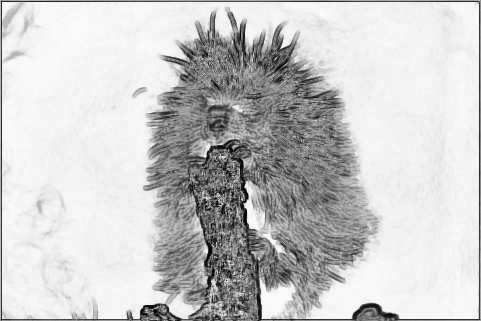}}
		&
		\fbox{\includegraphics[width=0.13\linewidth]{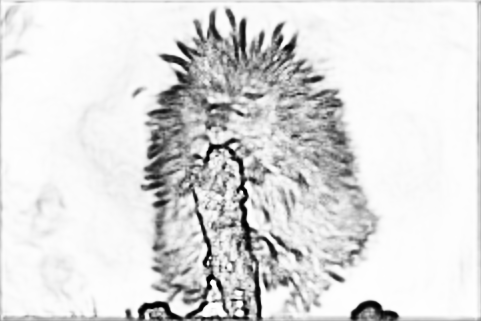}}
		&
		\fbox{\includegraphics[width=0.13\linewidth]{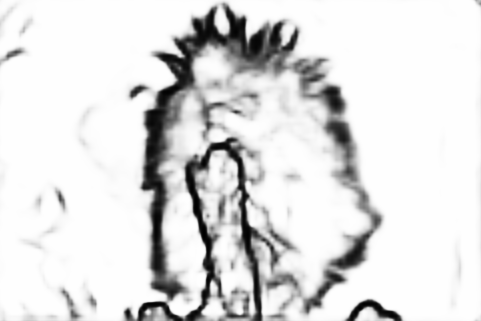}}
		&
		\fbox{\includegraphics[width=0.13\linewidth]{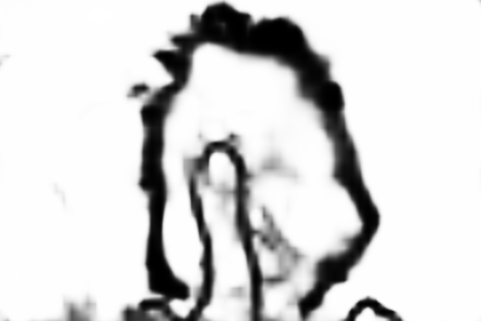}}
		&
		\fbox{\includegraphics[width=0.13\linewidth]{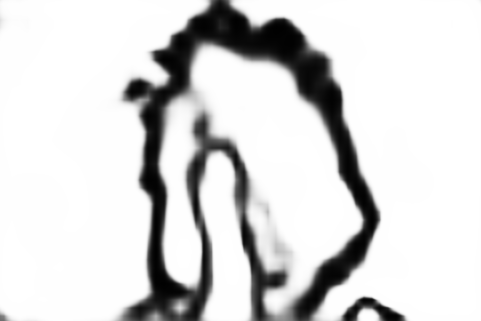}}\\
		\rotatebox{90}{CATS-RCF} &  \fbox{\includegraphics[width=0.13\linewidth]{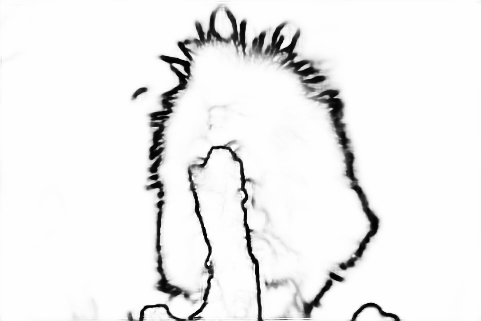}}
		& 
		\fbox{\includegraphics[width=0.13\linewidth]{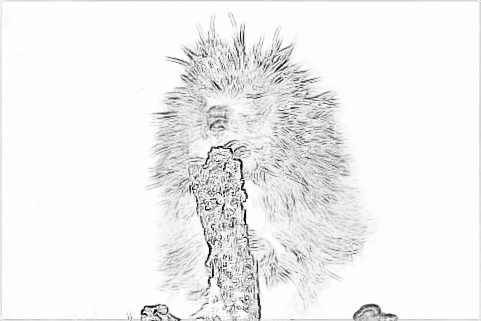}}
		&
		\fbox{\includegraphics[width=0.13\linewidth]{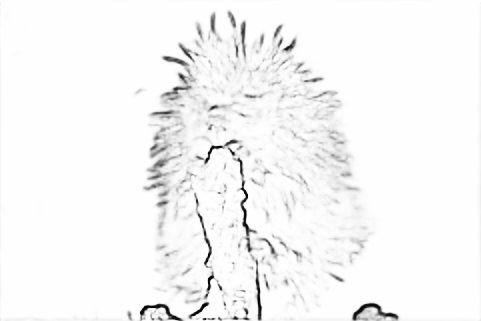}}
		&
		\fbox{\includegraphics[width=0.13\linewidth]{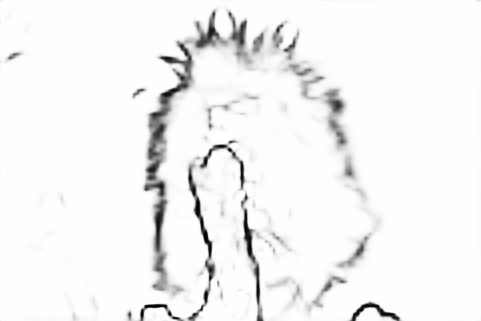}}
		&
		\fbox{\includegraphics[width=0.13\linewidth]{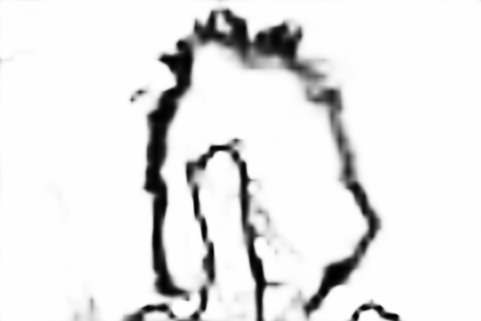}}
		&
		\fbox{\includegraphics[width=0.13\linewidth]{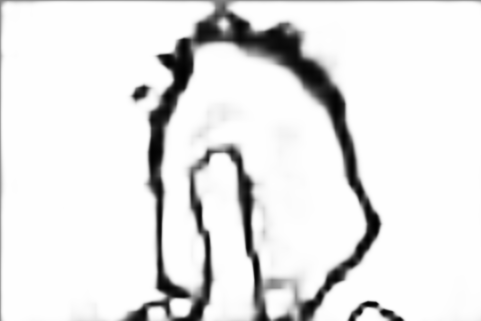}}
	\end{tabular}
	\vspace{-1mm}
	\caption{Qualitative comparison between the prior arts of deep edge detection (HED~\cite{xie2017hed} and RCF~\cite{liu2019richer}) and our proposed CATS. The top row displays an example image and its edge annotations from the BSDS500 dataset~\cite{amfm_pami2011}. The final edge predictions and the multi-level side edge maps estimated by different edge detection approaches are listed in rows. In the bottom line, we leverage our proposed CATS on RCF~\cite{liu2019richer} to obtain the most precisely-located side edge maps and final edge prediction.}
	\label{fig:teaser-image}
	\vspace{-4mm}
\end{figure*}

	Deep learning solutions dramatically improved the performance of edge detection as the classification on hierarchical deep features generated by large receptive fields can robustly suppress false alarms in texture regions~\cite{Yang2016Object, Maninis2018Convolutional, liu2016learning, he2019bi-directional, xie2017hed, liu2019richer}. These methods, however, have a challenging issue of detecting crisp edge maps that are free of localization ambiguity.	As shown in \figref{fig:teaser-image}, both the network outputs of HED~\cite{xie2017hed} and RCF~\cite{liu2019richer} suffer from localization ambiguity in the regions that contain true positive edge pixels across convolutional stages.
	To obtain crisp edge maps, it is required to use a morphological non-maximal suppression (NMS) scheme for the network outputs~\cite{Wei2015DeepContour, kokkinos2015pushing, Yang2016Object, xie2017hed, Ganin2014, liu2019richer, he2019bi-directional}.
	
	There have been some works that focused on the crispness of deep edge detection~\cite{Wang2018Deep, deng2018learning} without using the morphological NMS scheme.
	Wang \emph{et al.}~\cite{Wang2018Deep} proposed to use a new refinement architecture to tackle the problem of crispness. In the same period, Deng~\emph{et al.}~\cite{deng2018learning} found that the commonly-used weighted cross entropy loss is a key factor of posing the issue of localization ambiguity and adopted a Dice loss~\cite{milletari2016v} for edge detection. Limited by the numerical stability issue, the Dice loss tends to yield weaker edge response than the weighted cross entropy, which is disadvantageous for distinguishing edge and non-edge points, and~\cite{deng2018learning} thus has to add the ResNeXt block~\cite{xie2017aggregated} to obtain better features than the commonly-used vanilla VGG16~\cite{Simonyan2014Very} network.

	In this paper, we focus on the crispness of deep edge detection. That is to say, we expect the raw edge predictions by the convolution layers are with less localization ambiguity around the true positive predictions while maintaining strong capability of texture suppression. This is important for separating the adjacent edges that are glued together, as such glued edges in the raw predictions can be hard for the post-processing operation to handle. 
	In contrast to the previous works~\cite{deng2018learning, Wang2018Deep}, we are going to study if it is possible to learn crisp edge maps with the commonly-used backbone architecture VGG16~\cite{Simonyan2014Very} for edge detection. 
	
	We found that the issue of localization ambiguity for deep edge detectors is mainly caused by the mixing phenomenon in convolutional neural networks (CNNs).

    \begin{itemize}[leftmargin=2em]
 	\vspace{-2mm}
    \item[-] {\bf Feature Mixing.}
    Limited by the existence of max-pooling layers (for larger receptive field) and upsampling operators (for pixel-wise prediction), the higher-level convolutional features are spatially mixed.  
	When training a network to obtain a (higher-level) side edge map at a convolutional stage using the commonly-used weighted cross entropy loss or its variants, it is challenging to correctly classify multiple pixels sourced with the severe class-imbalance issue from the same mixed feature vectors.
	As a result, the higher convolutional stage is, the blurrier side edge maps are obtained by existing deep edge detectors . 
	\item[-] {\bf Side Mixing.} 
	When the side outputs are obtained, it is required to fuse them into a final edge prediction. The phenomenon of \emph{side mixing} is easily to be observed in existing detectors as they simply weight the side edge predictions according to the learned side importance. On the one hand, because the final edge prediction should be able to suppress the complicated textures in an image, the high-level side edges will have greater importance than low-level ones. On the other hand, such a fusion strategy equally treats all pixels in the same side output, thus making it hard to selectively preserve the complementary merits of different side edges.
	As shown in \figref{fig:teaser-image}, the final edge predictions of both HED and RCF are dominated by the high-level side edge maps while ignoring the low-level ones that have less localization ambiguities.
	\end{itemize}
		
	Based on the above discussion, we are going to unmix the convolutional features to learn crisp edge maps. First, we address the \emph{feature mixing} by tracing the true positive boundaries during training. We term the false positives caused by the \emph{feature mixing} as the \emph{confusing pixels} (of edges). To this end, we present a novel tracing loss with a specific focus on confusing pixels. We use a boundary tracing function to trace the actual position of an edge by enlarging the response difference between an edge pixel and the neighboring confusing pixels. Once the traced pixels are delineated, all the remained pixels belong to the background category and we only need to suppress the textures. Different from the existing loss functions for edge detection, we propose a texture suppression function to robustly handle texture regions by holistically suppressing the non-edge pixels that lie in the same image patch rather than treating these pixels as independent individuals. \textcolor{black}{For further alleviation of localization ambiguity, a pixel-wise fusion mechanism is required to handle the side mixing caused by image-level weighted average. We present a simple context-aware fusion (CoFusion) block that dynamically learns filters~\cite{DFN} with contextual information across side predictions under the general self-attention framework~\cite{xu2015show, chen2016attention, Woo_2018_ECCV}. A similar design of leveraging dynamic filters for edge perception task was previously presented in DFF~\cite{DFF}, it, however, focuses on fusing side predictions with high-level information for semantic edge detection. By contrast, our CoFusion block aims at addressing side mixing with the clues provided by the side edges themselves. The CoFusion block handles the side mixing that is hard for the tracing loss to alleviate with image-level weighted average; while the tracing loss offers necessary guidance to exert the CoFusion block.} 
	
	The tracing loss and the CoFusion block together form a context-aware tracing strategy (CATS) that is specifically designed for crisp edge detection.
	As displayed at the bottom of Fig.~\ref{fig:teaser-image}, with the proposed tracing function available, the side edge maps are reliably learned with less localization ambiguity, and the CoFusion fusion block makes the final edge predictions better in the aspect of crispness.

	In experiments, we leverage our proposed CATS on the existing state-of-the-art deep edge detectors HED~\cite{xie2017hed}, RCF~\cite{liu2019richer} and BDCN~\cite{he2019bi-directional} to demonstrate that the CATS can fully exploit the multi-level side outputs in terms of supervision and fusion to obtain precisely-located edge predictions. 
	In the standard evaluation of edge detectors, our proposed CATS consistently improves the accuracy of HED, RCF and BDCN on three common edge detection datasets. We achieve the state-of-the-art performance with a F-measure (ODS) of 0.812, 0.752, 0.897 on BSDS500~\cite{amfm_pami2011}, NYUDv2~\cite{Silberman:ECCV12} and MultiCue~\cite{M2016A} for RGB data, respectively.
	
	Furthermore, we quantitatively evaluate the aforementioned edge detectors with CATS on BSDS500 and NYUDv2 datasets without using the standard morphological non-maximal suppression scheme. The results validate that the CATS improves the F-measure (ODS) performance by large margins on BSDS500 and NYUDv2 with at most 12\% and 7.6\% respectively with robust edge localization ability.

	Our main contributions are summarized as follow: (1) We explicitly treat the false positive pixels as the confusing pixels, and propose a tracing loss to address the localization ambiguity for deep edge detectors. (2) We present a context-aware tracing strategy to learn precisely-located edge maps from natural images in an  end-to-end manner. (3) The proposed CATS consistently improves the performance of HED, RCF and BDCN on the BSDS500, the NYUDv2 and the Multicue datasets, and dramatically improves the edge localization accuracy of these detectors with no post-processing conditions.

	\section{Methodology}
	\label{sec:method}
	
	\subsection{Overview}
	\label{sec:overview}
	In this section, we elaborate the context-aware tracing strategy (CATS) that alleviates edge localization ambiguity by a \emph{tracing loss} and \emph{a context-aware fusion (CoFusion)} block for crisp edge learning.
	For feature unmixing with the weighted cross entropy, the tracing loss introduces a boundary tracing function to specifically separate the confusing pixels from edges, and a texture suppression function to perform holistic texture-region smoothing. Supervised by the tracing loss, the CoFusion block learns to selectively aggregate the different delineation merits of side edges in a pixel-wise fashion, which tackles the side mixing during the fusion. 
	
	\vspace{-1em}
	\subsection{The tracing loss}
	\subsubsection{Weighted cross entropy}
	Given an edge prediction $\hat{Y} = \{{\hat{y_i}}\}_{i = 1}^N$ and the corresponding edge label ${Y = \{{y_i}\}_{i = 1}^N}$, the weighted cross entropy is formulated as
	\begin{small}
		\begin{equation} \label{eq:bce}
			{L_{\text{ce}}}(\hat Y,Y) = -\lambda\,\alpha\sum_{i\in Y^{+}}{\log{\hat{y}_{i}}}-(1-\alpha)\sum_{i\in Y^{-}}{\log{(1-\hat{y}_{i})}},
		\end{equation}
	\end{small}where ${Y^+ = \{i|y_i\in Y, y_i > \delta\}}$, ${Y^- = \{i|y_i\in Y, y_i = 0\}}$ denote the edge and non-edge sample sets respectively, \textcolor{black}{and ${\alpha}$ is the proportion of the negative pixels in the set ${Y^+\cup Y^-}$.} The hyper-parameter ${\lambda}$ serves for importance balancing between edge and non-edge samples, and ${\delta}$ is a threshold to remove semantically controversial pixels when there are several annotators~\cite{liu2019richer, he2019bi-directional}.

\begin{figure}[t]
	\setlength{\fboxsep}{0pt}
	\begin{tabular}{cc}
		\fbox{\includegraphics[width=0.45\linewidth]{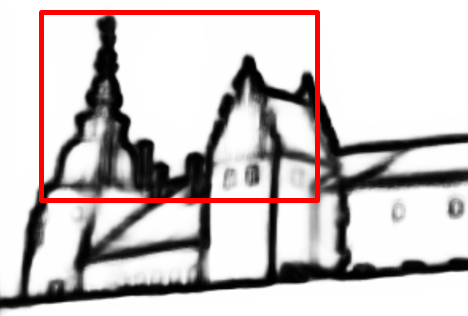}}&
		\fbox{\includegraphics[width=0.45\linewidth]{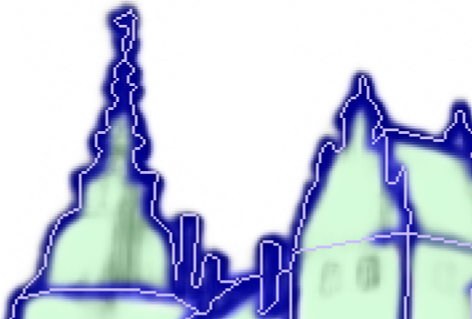}}\\
		\multicolumn{2}{c}{\includegraphics[width=0.87\linewidth]{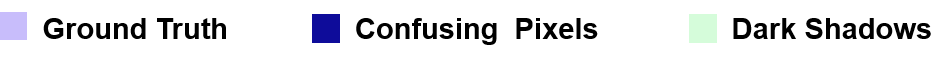}}
	\end{tabular}
	\vspace{-5mm}
	\caption{An illustration of confusing pixels. 
		The left image displays edges predicted by RCF~\cite{liu2019richer} and the right image is the enlarged rectangle region (red box) in the left image. Pixels in different colors in the right image indicate the ground truth edge pixels (purple), the confusing pixels of edges (blue), and the dark shadows in texture regions (light green).}
	\label{fig:boundary}
	\vspace{-3mm}
\end{figure}

	The weighted cross entropy can effectively supervise the network to learn reasonable edge maps, but it has highly imbalanced attention between edge and non-edge samples, making it hard to distinguish the confusing pixels that share features with edges and consistently smooth high-frequency regions. Consequently, as illustrated in \figref{fig:boundary}, the detected edge maps usually contain misclassified confusing pixels that incur thick edges around true positive edge pixels, and get false positives that form dark shadows in texture regions. 
	
	\subsubsection{Feature unmixing by tracing boundaries}
	Different from the non-edge pixels that lie beyond edges in a complete texture regions, the confusing pixels are intertwined with edges by the shared features. For a clear separation from edges, the confusing pixels therefore require an edge-aware suppression that is capable of feature unmixing in the high-level side edges, rather than being simply smoothed as texture points.

	Based on the above discussion, we devised a boundary tracing function to weaken the interaction between edges and their confusing neighbors by enlarging the response difference for crisp edge delineation. The boundary tracing function is formulated as 
	
	\begin{small}
		\begin{equation}\label{eq:bdr}
			L_{\text{bdry}}(\hat Y,Y) = - \sum\limits_{p \in E} \log
			\left(
			{\sum\limits_{i \in {L_p}} {{{\hat y}_i}} }/(\sum\limits_{i \in {R^{e}_p}\backslash{L_p}} {{{\hat y}_i}}  + \sum\limits_{i \in {L_p}} {{{\hat y}_i}})
			\right),
		\end{equation}
	\end{small}
	
	In \equref{eq:bdr}, ${E}$ is the set of all the edge points in the edge label ${Y}$. ${R^{e}_p}$ denotes a small image patch (e.g., a ${7 \times 7}$ rectangle patch) that contains edge fragments, and the patch centers at an edge point ${p}$. \textcolor{black}{The set of the edge points in ${R^{e}_p}$ is represented as ${L_{p}}$}. 
	
	Minimizing $L_{\text{bdry}}(\hat Y,Y)$ will force ${\sum_{i \in {R^{e}_p}\backslash{L_p}}{\hat y}_i}$ to be ${0}$ while increasing ${\sum_{i \in {L_p}}{{{\hat y}_i}}}$, which enlarges the response difference between edge fragments and accompanying confusing pixels. The boundary tracing function ${L_{\text{bdry}}}$ therefore smooths confusing pixels with awareness to the predicted response strength of neighboring edge fragments, and traces the actual edge position with less of the ambiguity caused by confusing pixels. \textcolor{black}{In practice, $L_{\text{bdry}}$ is computed via three convolution layers that are applied on the prediction and the label. The time complexity is about $\mathcal{O}((2+3k^2)\times M\times N)$ with an $M\times N$ image and a $k\times k$ patch size.}

	With edges and confusing pixels handled by the boundary tracing function, the remained texture regions can be consistently suppressed by a texture suppression function defined as
	\begin{equation}
		{{L_{\text{tex}}}(\hat Y,Y) =  - \sum\limits_{p \in Y\backslash\hat E} 
			\log (1 - {\sum\limits_{i \in {R^{t}_p}} {\hat y}_i }/{|{R^{t}_p}|}
			)
			,}
		\label{eq:tex}
	\end{equation}
	where ${R^{t}_p}$ represents an image patch that centers at a non-edge point $p$ (e.g., a ${3 \times 3}$ rectangle area), and ${\hat E}$ is a set that includes all edges and their confusing pixels that have been used in the boundary tracing function. ${\hat E}$ serves as a buffer zone to weaken the negative interaction between edge fragments and texture areas that require a stronger suppression than confusing pixels.

	As revealed by \equref{eq:tex}, the texture suppression function groups the non-edge points in the same texture region patch, and \textcolor{black}{holistically suppresses these pixels} rather than treating them as independent individuals. Focusing on complete texture regions beyond edges, the texture suppression function actually works in a complementary way with the boundary tracing function.

	With the boundary tracing function and the texture suppression function, the tracing loss follows as 
	\begin{equation}
		{\text{TracingLoss}(\hat Y,Y) = {L_{\text{ce}}} + {\lambda _1}{L_{\text{bdry}}} + {\lambda _2}{L_{\text{tex}}}},
		\label{eq:CATS}
	\end{equation}
	where $\hat Y$ and $Y$ respectively denote the edge prediction and the edge label, $\lambda_{1}$ and $\lambda_{2}$ are hyper-parameters to balance each element in the tracing loss. 
	
	During model training, ${L_{\text{ce}}}$ performs a rough edge learning, ${L_{\text{bdry}}}$ tackles edge localization refinement by feature unmixing, and ${L_{\text{tex}}}$ imposes a strong holistic suppression on texture regions. With ${L_\text{{bdry}}}$ and ${L_\text{{tex}}}$, the tracing loss processes the non-edge points that are gathered according to their surrounding context with target-specific suppression, and achieves crisp edge generation with less localization ambiguity than single weighted cross entropy.

	\subsection{Context-aware fusion block}
	To leverage the edge details in low-level edge maps and the global context with robust texture suppression in the high-level counterparts, weighted average is a common operation adopted in previous works for obtaining a unified edge prediction from multi-level side edges~\cite{xie2017hed, liu2019richer, he2019bi-directional}. Albeit effective, this average operation incurs the issue of side mixing as all the pixels in a side edge map share the same fixed weight and have equal importance during fusion.

\begin{figure}[t!]
	\centering
	\includegraphics[width=\linewidth,keepaspectratio]{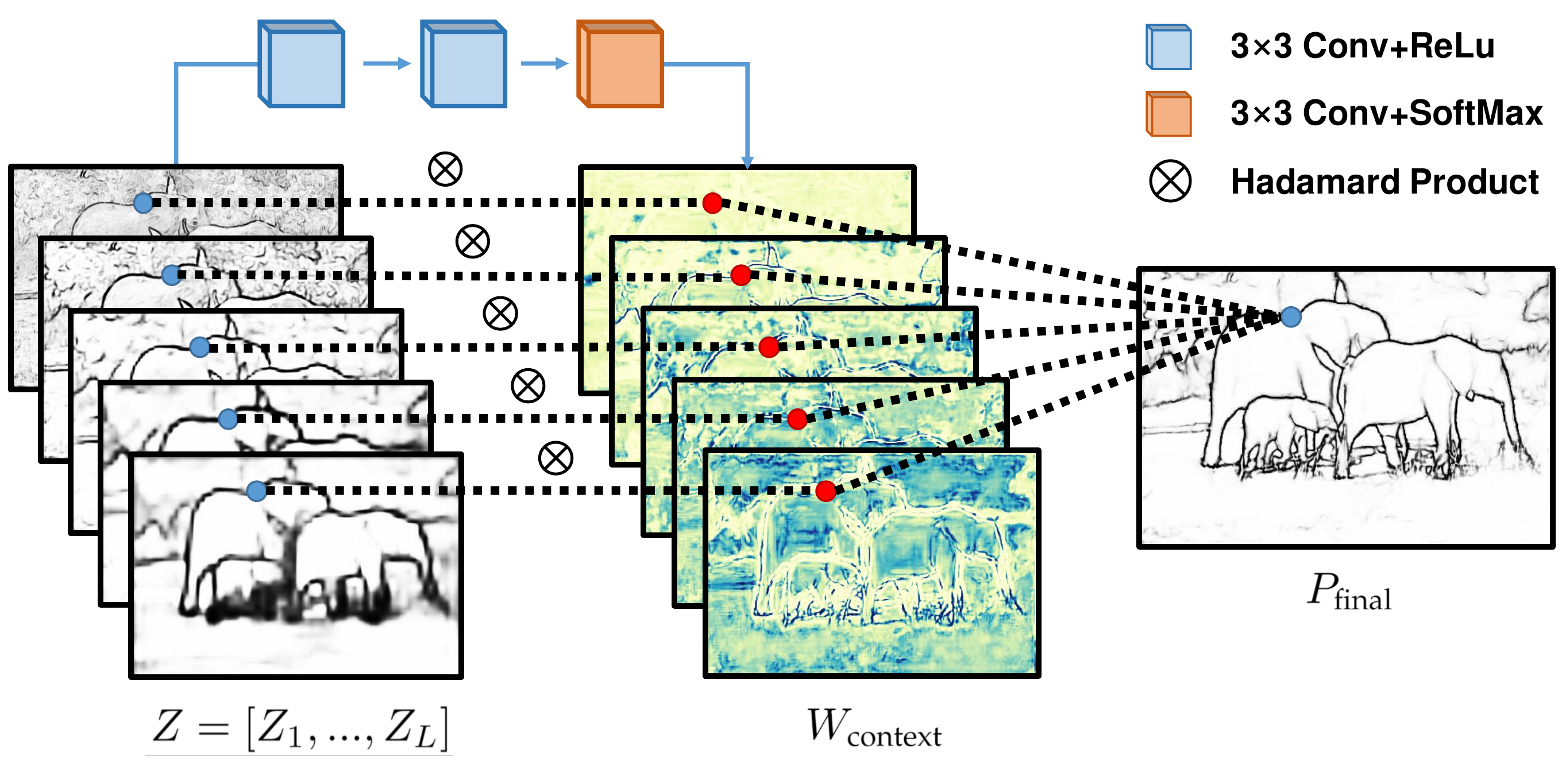}
	\vspace{-9mm}
	\caption{The mechanism of the CoFusion Block.}
	\vspace{-3mm}
	\label{fig:CATSFusing}
	\end{figure}

	For harnessing the multi-level side maps without side mixing, as shown in \figref{fig:CATSFusing}, we designed a context-aware fusion (CoFusion) block within a self-attention framework to absorb the merits and bypass the limitations of different side maps by a position-wise aggregation of side edges.

	Let ${Z = [{Z_1},...,{Z_L}] \in \mathbb{R}^{H \times W \times L}}$ denote $L$ side edge heatmaps. The CoFusion block first learns a weight map ${W_{\text{context}} \in \mathbb{R}^{H \times W \times L}}$ from ${Z}$ by an attention block. To derive a score map ${A_{\text{score}} = [a_{ijl}] \in \mathbb{R}^{H \times W \times L}}$ from ${Z}$, the attention block adopts three $3 \times 3$ convolution layers to capture contextual information for inferring $W_{\text{context}}$. ${A_{\text{score}}}$ is subsequently normalized through softmax activation to get a weight map ${W_{\text{context}} = [w_{ijl}]}$, where  
 
	\begin{align}
		w_{ijl} = e^{a_{ijl}}/\sum_{k = 1}^L e^{a_{ijk}}.
	\end{align}
	A vector ${{v_{ij}} \in {\mathbb{R}^{1 \times 1 \times L}}}$ in ${{W_{\text{context}}}}$  determines how much each side map contributes to the pixel ${p_{ij}}$ of the final edge prediction ${P_{\text{final}}}$. 
	Because ${v_{ij}}$ is calculated with the features around ${p_{ij}}$ across side maps, ${{W_{\text{context}}}}$ changes according to the context information in the multi-level side maps and leverages their different delineation preferences for the final prediction. With the context-aware weights ${{W_{\text{context}}}}$, ${P_{\text{final}}}$ can be calculated as follows.
	\begin{align}
		&{\text{CoFusion}(Z) = \sum\limits_{l = 1}^L {{W_{\text{context}}} \otimes Z}} ; \\ 
		&P_{\text{final}} = \text{sigmoid}(\text{CoFusion}(Z)), 
	\end{align}
	where ${\otimes}$ denotes Hadamard product. 

	\section{Experiments and analysis}
	\label{sec:experiment}
	To validate the effectiveness and generality of the CATS, we implement it in Pytorch framework~\cite{steiner2019pytorch:} with three VGG16-based~\cite{Simonyan2014Very} edge detectors, including HED~\cite{xie2017hed}, RCF~\cite{liu2019richer} and BDCN~\cite{he2019bi-directional}. In this section, we evaluate the CATS on three challenging edge detection benchmarks, including NYUDv2~\cite{Silberman:ECCV12}, BSDS500~\cite{amfm_pami2011} and Multicue~\cite{M2016A} datasets. An ablation study is subsequently presented for a more comprehensive performance analysis of the CATS.
	\subsection{Datasets}
	\paragraph*{BSDS500~\cite{amfm_pami2011}} This dataset is a challenging edge detection benchmark that is composed of 200 training, 100 validation and 200 test images. Each image in this dataset is annotated by several annotators. The training and validation data are jointly augmented for model training as in \cite{xie2017hed} and \cite{liu2019richer}.
	
	\paragraph*{NYUDv2~\cite{Silberman:ECCV12}} This is a challenging dataset for indoor scene parsing and is also a commonly used benchmark for edge detection evaluation~\cite{xie2017hed, deng2018learning, liu2019richer, he2019bi-directional}. It contains 1449 densely annotated RGB-D images, and is divided into 381 training, 414 validation and 654 testing images. We use this dataset to compare our proposed CATS with the state-of-the-art edge detectors, and also perform a comprehensive analysis on the NYUDv2 dataset in the ablation study.

	\paragraph*{Multicue dataset~\cite{M2016A}} This dataset strictly distinguishes the definitions of boundary and edge, and it thus consists of two sub datasets: Multicue Boundary and Multicue Edge. This dataset regards semantically meaningful pixels as object boundary points, and treats pixels with abrupt perceptional changes as low-level edge points. We testify the CATS on both sub-datasets. Same with~\cite{M2016A, xie2017hed, liu2019richer, he2019bi-directional}, we randomly split the Multicue dataset into 80 training and 20 testing samples, and conducted three independent trials on both sub-datasets.

	\vspace{-1em}
	\subsection{Implementation details}
	Experiments in this paper are conducted on a single GeForce RTX 2080 Ti GPU. The model parameters are updated by stochastic gradient descent (SGD) optimizer with momentum 0.9 and weight decay 0.0002, and the batch size is set to 10. The initial learning rate is 1e-6 and is multiplied with 0.1 after every given period shown in \tabref{tab:params}. The hyper-parameters in the tracing loss are also specified in \tabref{tab:params}. 
	
	\begin{table}[!ht]
		\vspace{-3mm}
		\centering
		\caption{Parameter setting of the tracing loss in experiments.}
		\label{tab:params}
		\vspace{-0.1in}
		\scriptsize
		\begin{tabular}{p{6em}<{\centering}|c|c|c|c|c|c} \hline
			\multirow{2}{*}{\diagbox[width=7.7em]{Data}{Param.}}
			& \multirow{2}{*}{period/epoch} & \multirow{2}{*}{${\delta}$} & \multirow{2}{*}{${\lambda}$} & \multicolumn{3}{c}{${\lambda_{1}/{\lambda_{2}}}$} \\
			\cline{5-7}
			& & & & L1-L3 & L4-L5 & Final \\
			\hline
			NYUDv2 & 20/60 & - & 1.2 & 4/0.05 & 2/0.1 & 6/0.05 \\
			BSDS500 & 10/40 & 0.3 & 1.1 & 2/0.05 & 1/0.1 & 4/0.05 \\
			Multicue\_bdr & 20/60 & 0.3 & 1.2 & 2/0.05 & 1/0.1 & 4/0.03 \\
			Multicue\_edge & 20/60 & 0.2 & 1.1 & 4/0.01 & 2/0.01 & 6/0.01 \\ 
			
			\hline
		\end{tabular}
		\vspace{-3mm}
	\end{table}
	
	\begin{figure*}[!t]
	\setlength{\fboxsep}{0pt}
	\begin{tabular}{cccccc}
		\fbox{\includegraphics[width=0.14\linewidth]{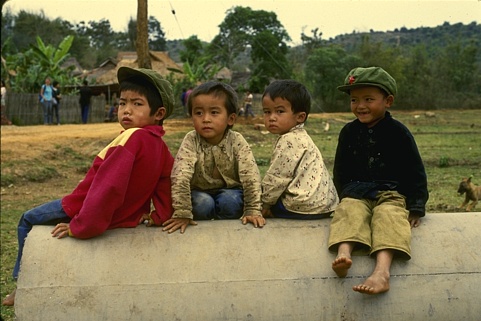}}&
		\fbox{\includegraphics[width=0.14\linewidth]{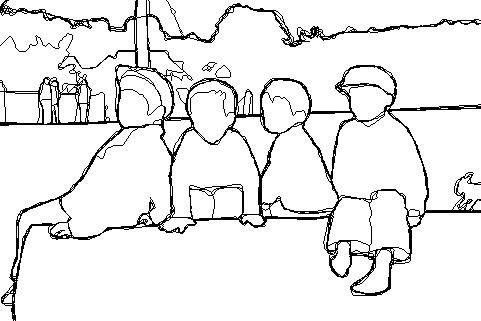}}&
		\fbox{\includegraphics[width=0.14\linewidth]{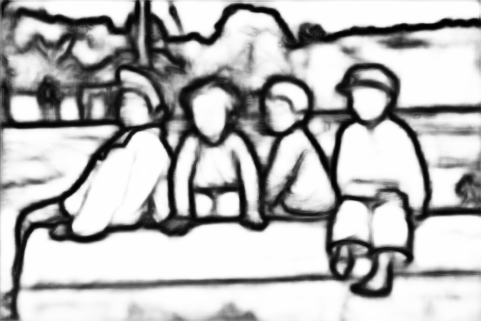}}&
		\fbox{\includegraphics[width=0.14\linewidth]{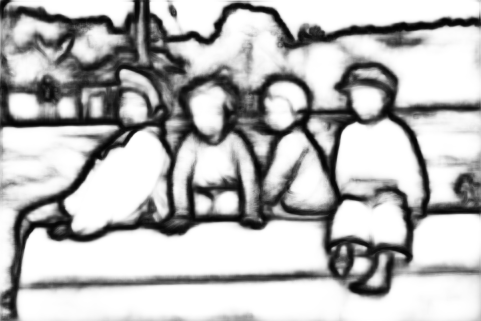}}&
		\fbox{\includegraphics[width=0.14\linewidth]{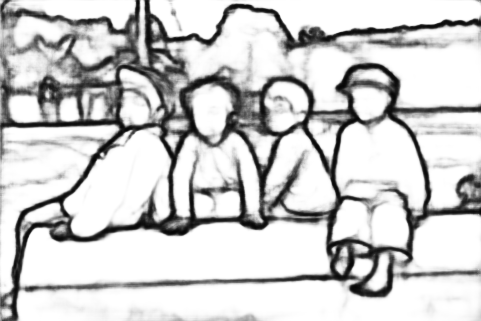}}&
		\fbox{\includegraphics[width=0.14\linewidth]{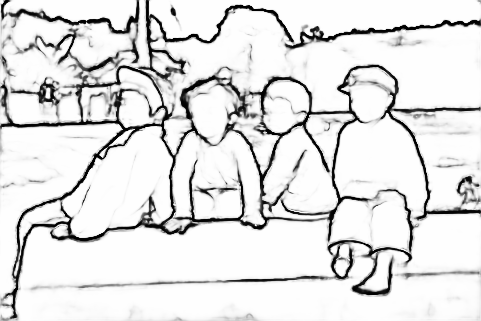}}
		\\
		
		\fbox{\includegraphics[width=0.14\linewidth]{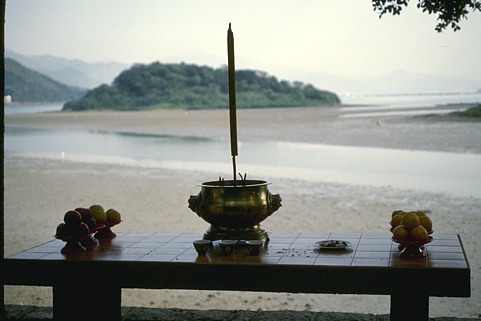}}&
		\fbox{\includegraphics[width=0.14\linewidth]{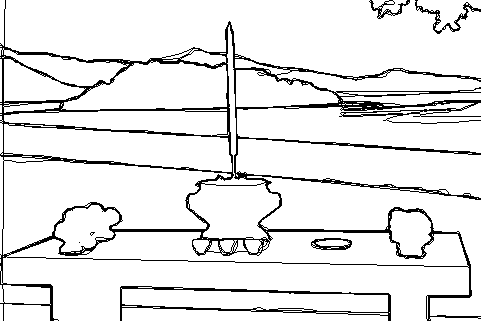}}&
		\fbox{\includegraphics[width=0.14\linewidth]{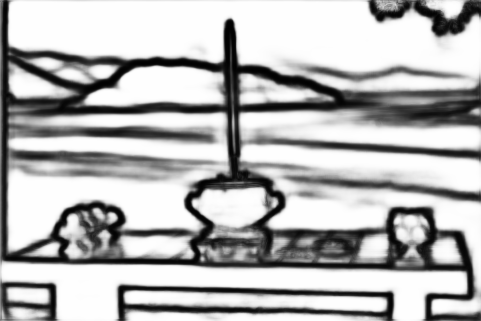}}&
		\fbox{\includegraphics[width=0.14\linewidth]{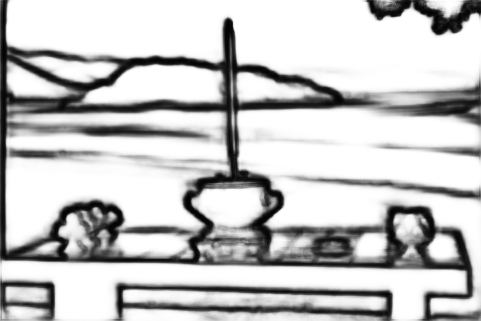}}&
		\fbox{\includegraphics[width=0.14\linewidth]{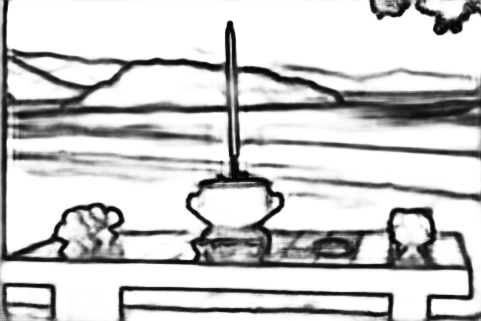}}&
		\fbox{\includegraphics[width=0.14\linewidth]{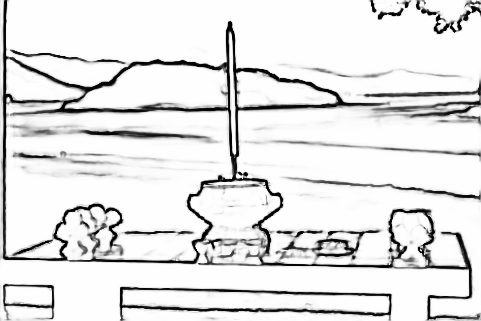}}
		\\
		
		\fbox{\includegraphics[width=0.14\linewidth]{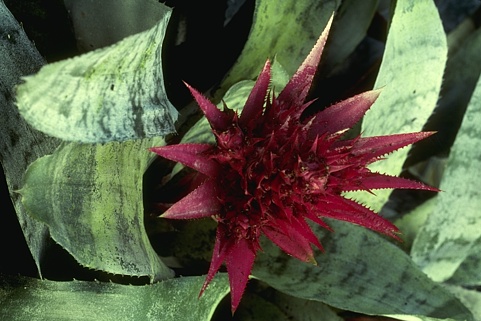}}&
		\fbox{\includegraphics[width=0.14\linewidth]{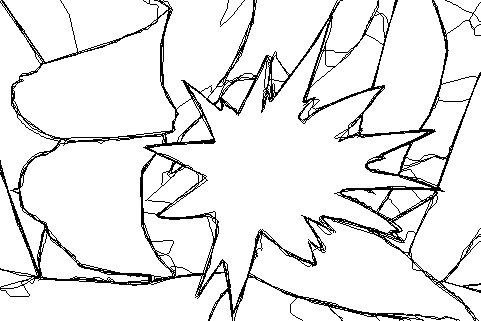}}&
		\fbox{\includegraphics[width=0.14\linewidth]{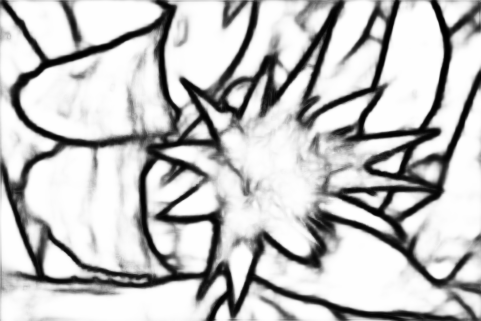}}&
		\fbox{\includegraphics[width=0.14\linewidth]{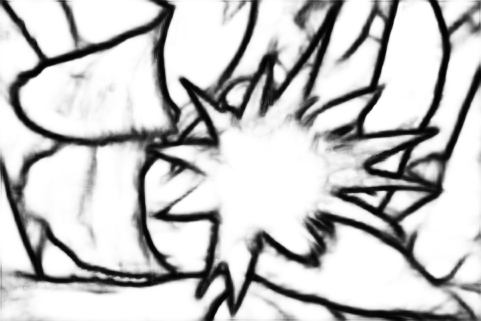}}&
		\fbox{\includegraphics[width=0.14\linewidth]{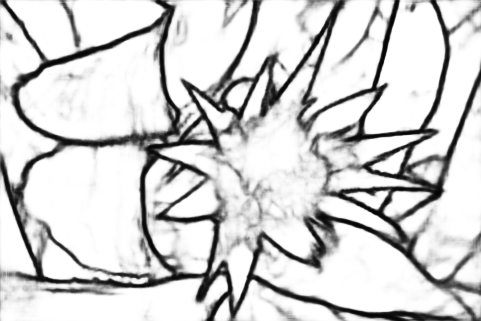}}&
		\fbox{\includegraphics[width=0.14\linewidth]{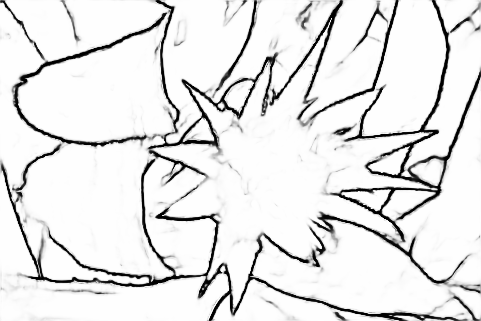}}
		\\
		
		\fbox{\includegraphics[width=0.14\linewidth]{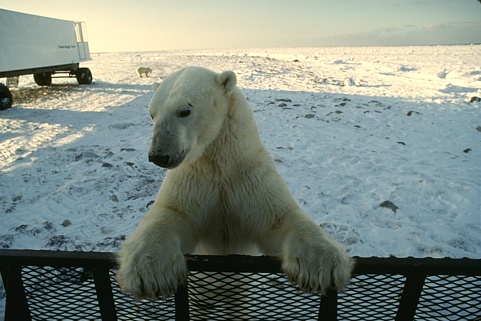}}&
		\fbox{\includegraphics[width=0.14\linewidth]{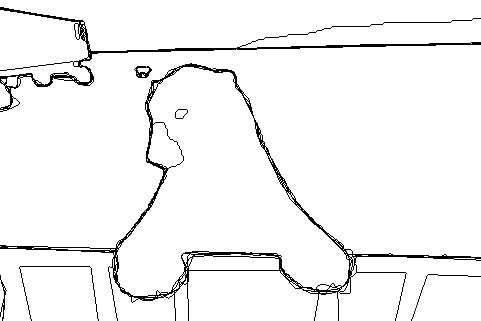}}&
		\fbox{\includegraphics[width=0.14\linewidth]{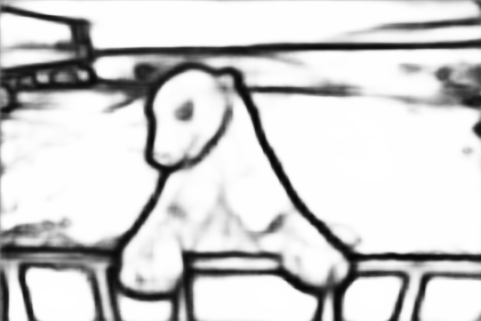}}&
		\fbox{\includegraphics[width=0.14\linewidth]{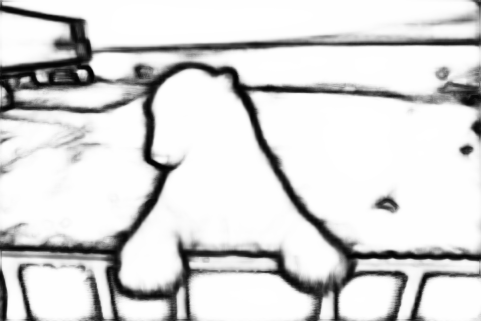}}&
		\fbox{\includegraphics[width=0.14\linewidth]{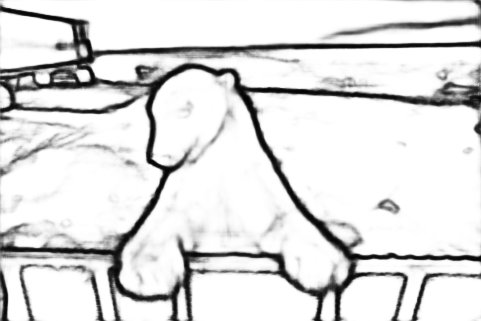}}&
		\fbox{\includegraphics[width=0.14\linewidth]{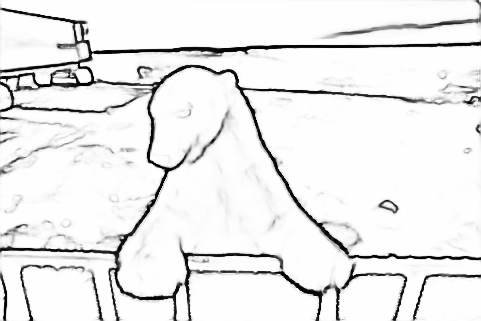}}
		\\
		Image & Label & HED~\cite{xie2017hed} & RCF~\cite{liu2019richer} & BDCN~\cite{he2019bi-directional} & CATS-RCF

	\end{tabular}
	\caption{Qualitative comparison between the prior arts of deep edge detection and our proposed CATS on the BSDS500 dataset~\cite{amfm_pami2011}.}
	\vspace{-4mm}
	\label{fig:vis}
\end{figure*}
	
	\subsection{Evaluation protocols}
	The core step of evaluating an edge detector is to pixel-wisely match the ground truth of edges to the binarized edge prediction under a specified maximum allowed distance tolerance. Following the previous works, a parametric curve of the precision and recall will be drawn to evaluate the overall performance of an edge detector with different thresholds of binarization. The best performance of an edge detector will be reported by the F-measure scores with the optimal thresholds at both dataset scale (ODS) and image scale (OIS). The maximum allowed distance tolerance for correct matches between the edge predictions and the annotations is conventionally set to 0.0075 for the BSDS500 and Multicue datasets, and 0.011 for the NYUDv2 dataset~\cite{xie2017hed, liu2019richer, he2019bi-directional, Maninis2018Convolutional}.
	
	\paragraph*{Standard evaluation protocol} Before binarizing a given edge prediction, the general evaluation procedure in previous researches will first apply a standard post-processing scheme. The post-processing scheme includes a non-maximum suppression (NMS) step and a mathematical morphology operation to obtain a thinned edge map, and the standard evaluation protocol then uses the thinned edge prediction for matching with the ground truth of edge maps. 
	
	\paragraph*{Crispness-emphasized evaluation protocol} Although the post-processing scheme can partially remove the falsely-alarmed edge pixels around the true positive ones, it would be interesting to explore whether the raw outputs of the deep edge detectors have a better localization ability of edges. Accordingly, we remove the standard post-processing scheme to evaluate the performance of a trained edge detector in the aspect of crispness. 
	
	\begin{figure*}[!t]
	\begin{tabular}{c}
		
		\hspace{-0.15in}\subfloat[]{\includegraphics[width=0.25\linewidth]{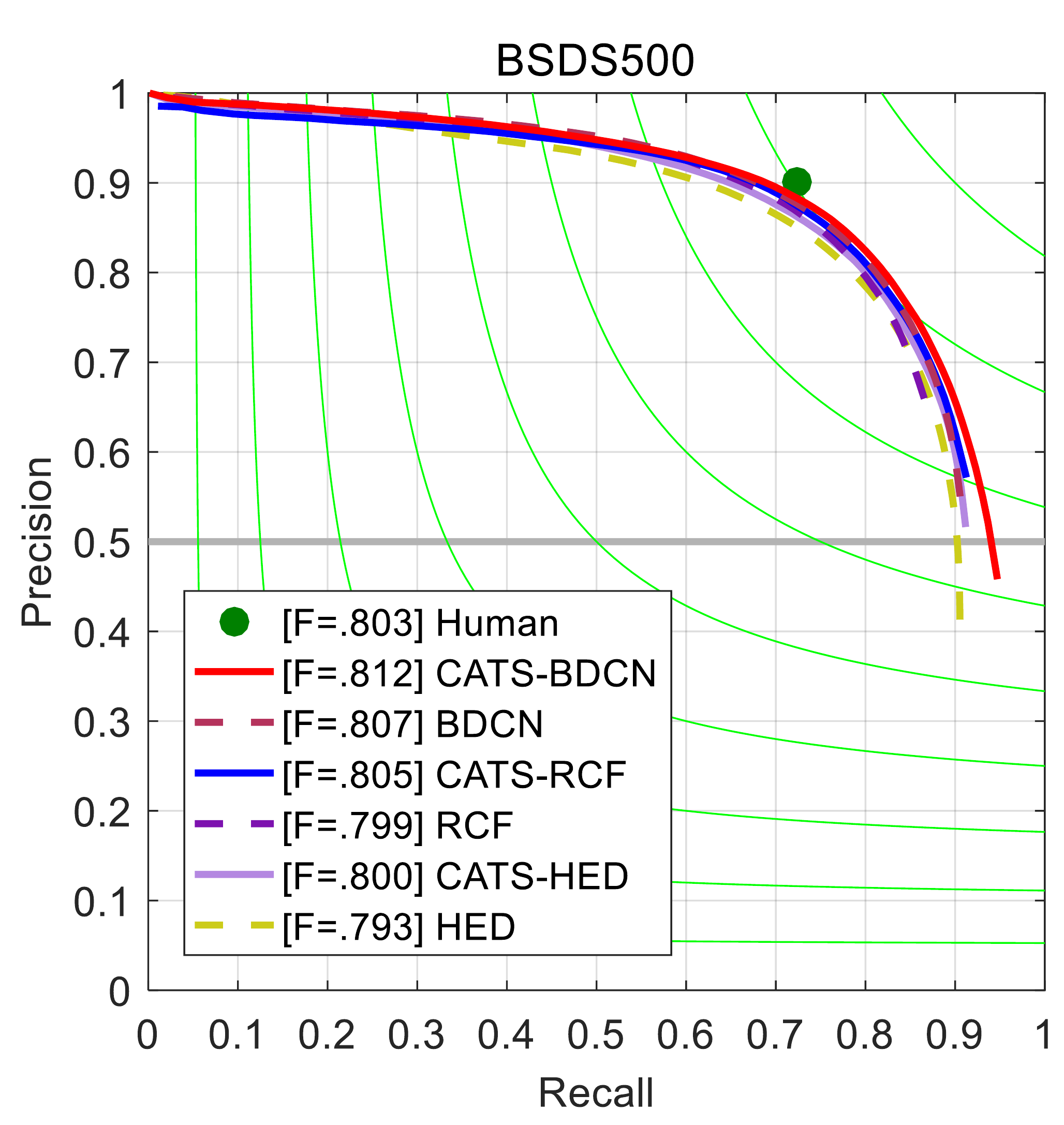}}%
		
		\subfloat[]{\includegraphics[width=0.25\linewidth]{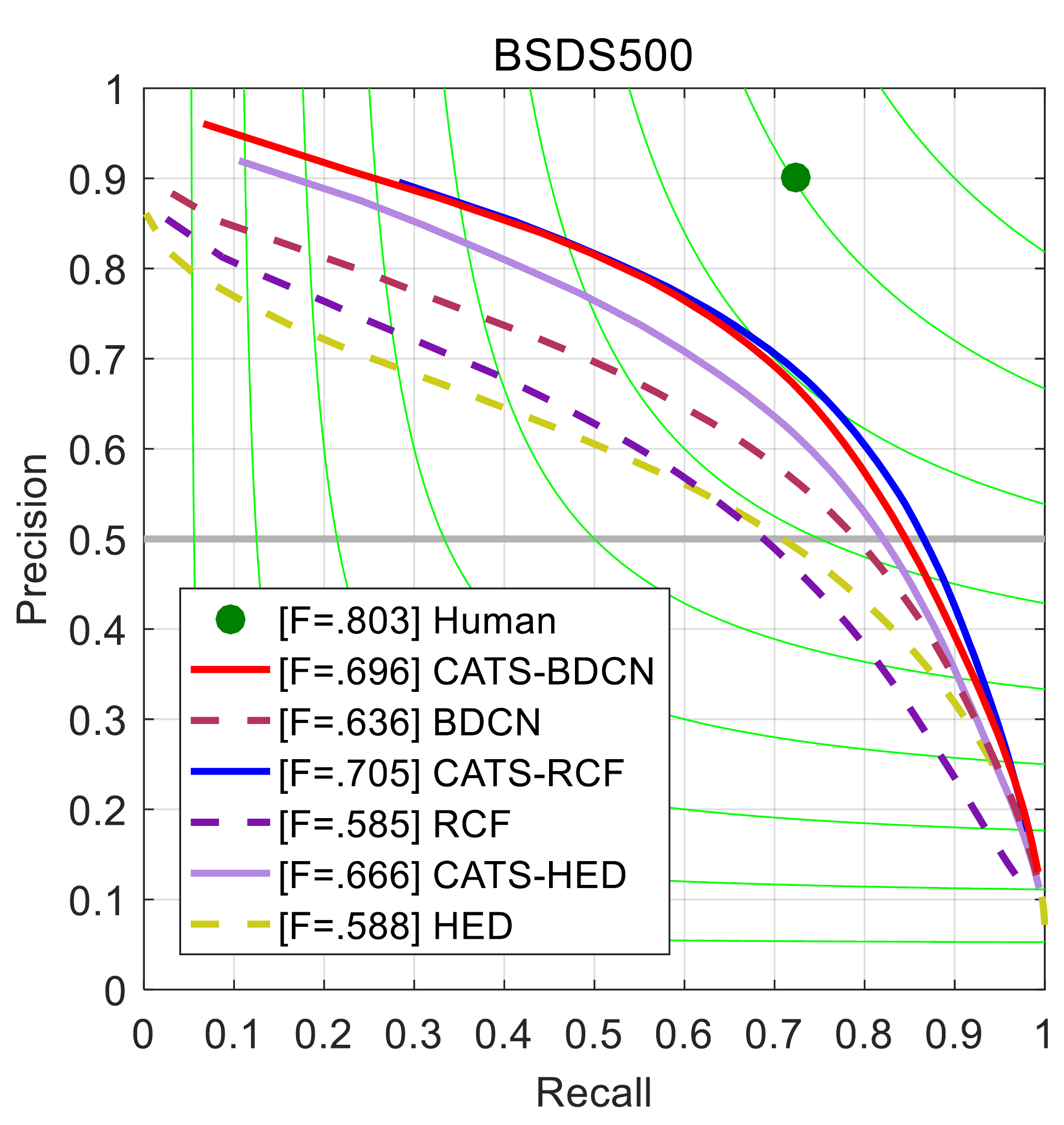}}%
		
		\subfloat[]{\includegraphics[width=0.25\linewidth]{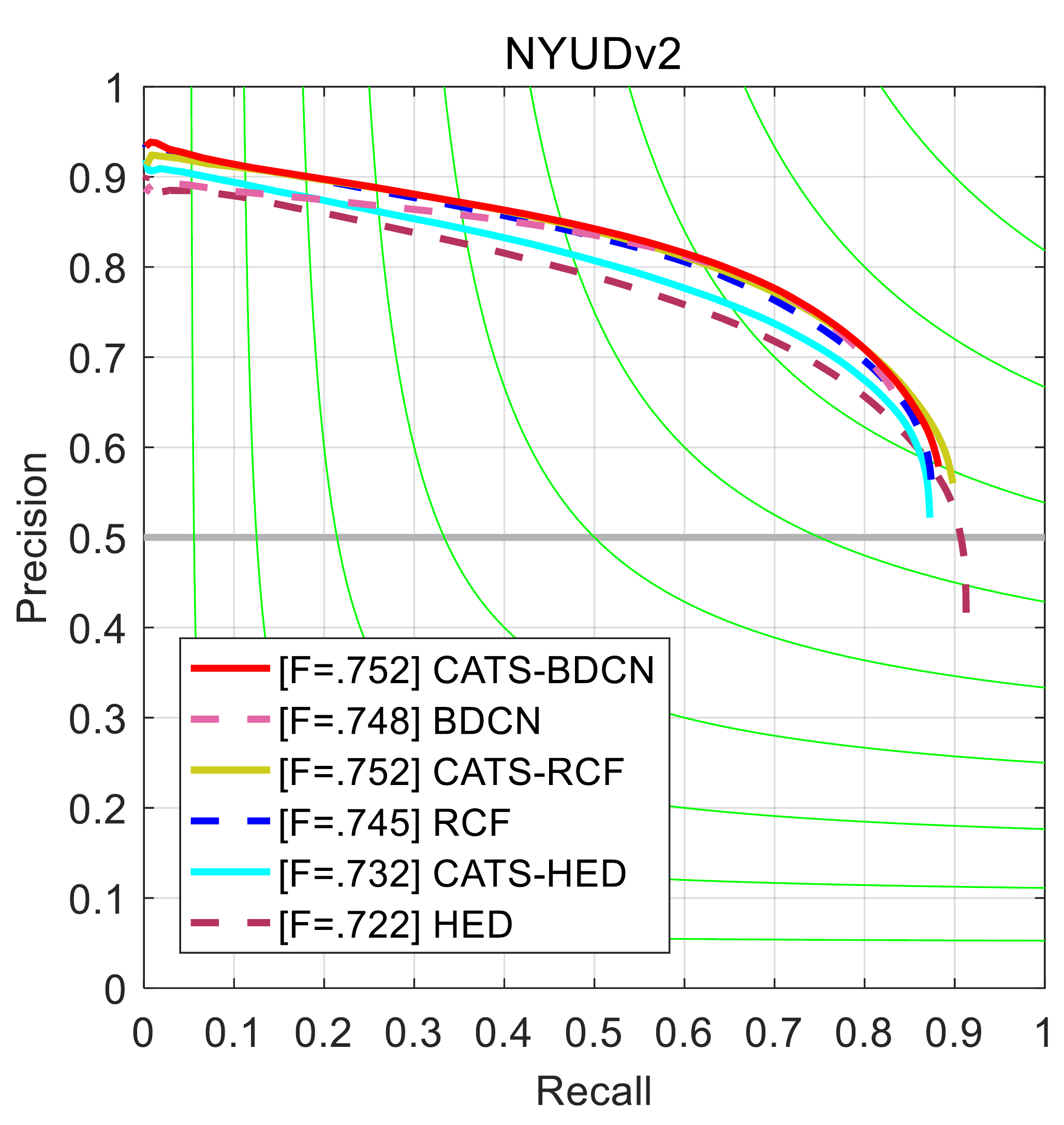}}%
		
		\subfloat[]{\includegraphics[width=0.25\linewidth]{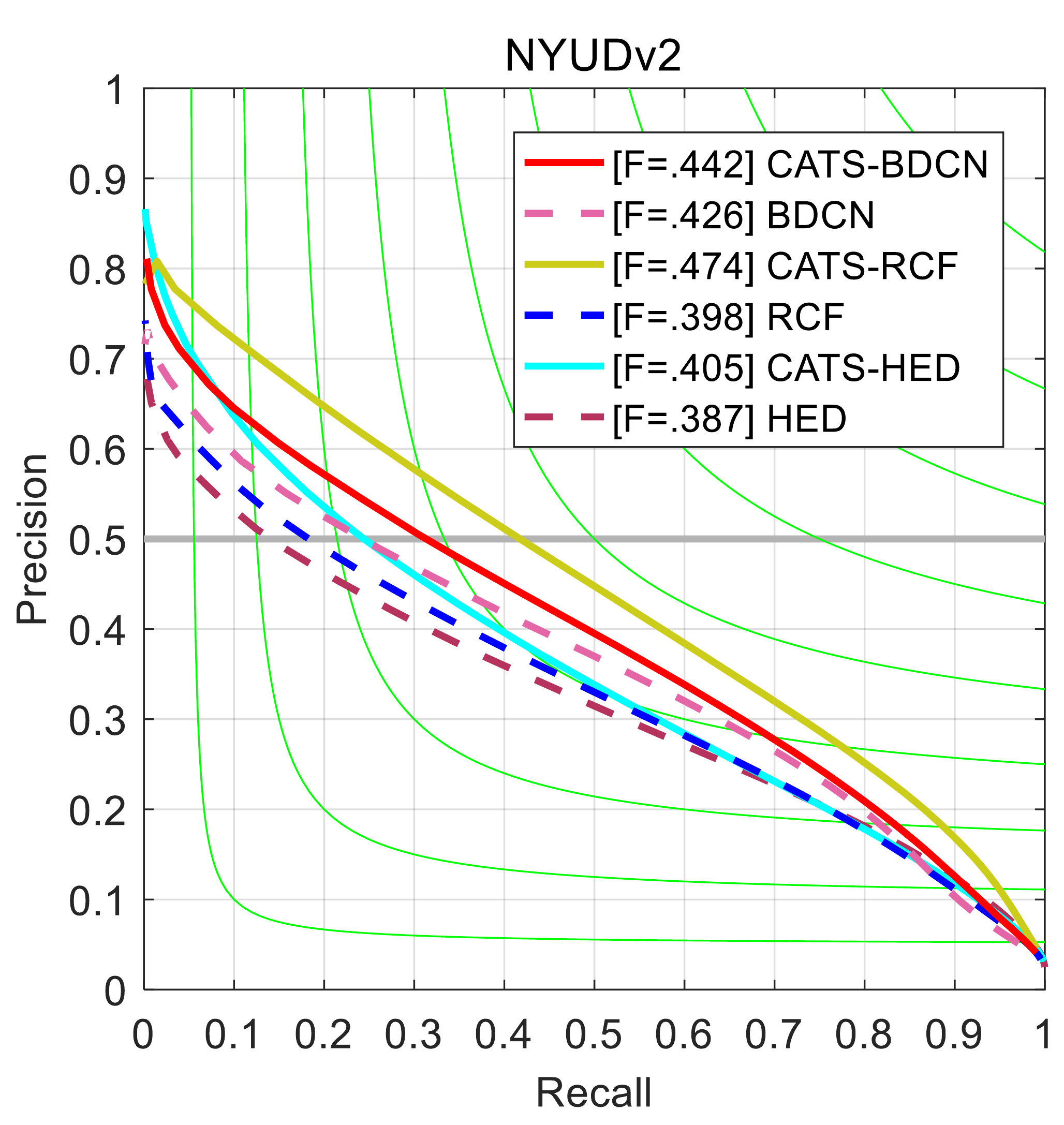}}%

	\end{tabular}
	\vspace{-4mm}
	\caption{The precision-recall curves of the CATS-based and the compared original models on BSDS500 and NYUDv2 datasets. (a) and (c) depict the results under the standard evaluation, while (b) and (d) are the results of the crispness-emphasized evaluation.}
	\label{fig:pr}
	\vspace{-4mm}
\end{figure*}
	
	\subsection{Comparison with state-of-the-art methods}
	In this section, we report the statistic comparison of our method with existing state-of-the-art models on BSDS500, NYUDv2 and Multicue datasets. We tested and validated the proposed CATS on the VGG16-based HED~\cite{xie2017hed}, RCF~\cite{liu2019richer} and BDCN~\cite{he2019bi-directional}. Models in experiments were fine-tuned with a VGG16~\cite{Simonyan2014Very} model pre-trained on ImageNet~\cite{deng2009imagenet}. 
	
	\subsubsection{BSDS500 Dataset}
	The numerical results on BSDS500 dataset are listed in \tabref{tab:bsds_study}, showing that the application of CATS achieves consistent performance boosting with HED, RCF and BDCN, by at most 0.7\% in both ODS and OIS under the standard evaluation protocol.
	
	\begin{table}[!ht]
		\vspace{-3mm}
		\centering
		\caption{Quantitative analysis on the BSDS500 dataset. SEval denotes the standard evaluation, and CEval is the crispness-emphasized evaluation. For fair comparison, we only list the single-scale results generated by models trained with only BSDS500 data.}
		\label{tab:bsds_study}
		\vspace{-0.1in}
		\begin{tabular}{c||c|c|c|c}
			\hline
			\multirow{2}{*}{Methods} & \multicolumn{2}{c|}{SEval} & \multicolumn{2}{c}{CEval}\\ 
			\cline{2-5}
			& ODS & OIS & ODS & OIS\\
			\hline
			Human & 0.803 & 0.803 & - & -\\
			OEF\cite{Hallman2015Oriented} & 0.746	& 0.770 & - & -\\
			N\textsuperscript{4}-Fields\cite{Ganin2014} & 0.753 & 0.769 & - & -\\
			DeepContour\cite{Wei2015DeepContour} & 0.757 & 0.776 & - & -\\
			HFL\cite{bertasius2015HFL} & 0.767 & 0.788 & - & -\\
			CEDN\cite{Yang2016Object} &0.788 & 0.804 & - & -\\
			DeepBoundary\cite{kokkinos2015pushing} & 0.789 & 0.811 & - & -\\
			COB\cite{Maninis2018Convolutional} & 0.793 & 0.820 & - & -\\
			CED\cite{Wang2018Deep} & 0.794 & 0.811 & \textcolor{black}{0.642} & \textcolor{black}{0.656}\\
			AMH-Net\cite{xu2017learning} & 0.798 & 0.829 & - & -\\
			DCD\cite{liao2017deep-learning-based} & 0.799	& 0.817 & - & -\\
			LPCB\cite{deng2018learning} & 0.800 & 0.816 & \textcolor{black}{0.693} & \textcolor{black}{0.700}\\
			\hline
			HED\cite{xie2017hed} & 0.793 & 0.811 & 0.588 & 0.608\\
			CATS-HED & \textbf{0.800 }&\textbf{ 0.816} &  \textbf{0.666} & \textbf{0.676}\\
			RCF\cite{liu2019richer} & 0.799 &0.815 & 0.585 & 0.604 \\
			CATS-RCF & \textbf{0.805} &\textbf{ 0.822} & \textbf{0.705} & \textbf{0.716} \\
			BDCN\cite{he2019bi-directional} & 0.807 & 0.822 & 0.636  & 0.650 \\	
			CATS-BDCN & \textbf{0.812} & \textbf{0.828} & \textbf{0.696} & \textbf{0.705}\\
			\hline
		\end{tabular}
		\vspace{-1mm}
	\end{table}
	In \tabref{tab:bsds_study}, the CATS-applied models gain over the original counterparts by a dramatic margin in terms of the ODS and OIS scores of crispness-emphasized evaluation. The significant improvement indicates that the CATS can facilitate accurate edge localization with a robust removal of false positives along edges (\ie the confusing pixels). One may notice that the standard evaluation scores do not always synchronize with the crispness-emphasized evaluation counterparts, although they show simultaneous performance improvement when using the CATS on the same model. The post-processing operation will help mitigate the impact of the false positives around true positive edges, and the standard evaluation is therefore more influenced by the misclassified texture points than the confusing pixels. Despite of the different focuses of the two evaluation protocols, the results still indicate that the standard evaluation can benefit from less localization ambiguity.

	Visualization results on BSDS500 dataset can be found in \figref{fig:vis}. In the given examples, the CATS successfully tackles the confusing pixels caused by the issue of feature mixing and side mixing during fusion to generate crisper edges than other networks, which in turn makes adjacent boundaries easier to distinguish. More detailed comparison is supplied in \figref{fig:pr}, which gives the precision-recall curves of the CATS-based models and the corresponding originals on BSDS500 regarding the two evaluation schemes.
	
	\subsubsection{NYUDv2 Dataset}
	We independently conducted experiments on the RGB and HHA data provided in the NYUDv2 dataset, and averaged the RGB- and HHA-based predictions to generate merged edge predictions like previous work~\cite{xie2017hed, liu2019richer, he2019bi-directional}. The quantitative comparison with several existing methods is reported in \tabref{tab:nyud_study}.
	
	\begin{table}[!h]
		\vspace{-3mm}
		\centering
		\caption{Quantitative analysis on the NYUDv2 dataset. SEval and CEval respectively refer to the standard evaluation and the crispness-emphasized evaluation.}
		\label{tab:nyud_study}
		\vspace{-0.1in}
		\begin{tabular}{c||c|c|c|c|c}
			\hline
			\multirow{2}{*}{Methods} & \multirow{2}{*}{Data} & \multicolumn{2}{c|}{SEval} & \multicolumn{2}{c}{CEval}\\ 
			\cline{3-6}
			& & ODS & OIS & ODS & OIS\\
			\hline
			\multirow{3}{*}{LPCB\cite{deng2018learning}} & RGB &0.739 &0.754 & - & - \\
			&HHA&0.707&0.719 & - & - \\
			&RGB-HHA& 0.762 & 0.778 & - & - \\
			\hline
			\multirow{3}{*}{HED\cite{xie2017hed}} & RGB &0.722 &0.737 & 0.387 & 0.404\\
			&HHA&0.691 &0.704 & 0.335 & 0.350\\
			&RGB-HHA& 0.746 & 0.764 &  0.368 & 0.384\\
			\hline
			\multirow{3}{*}{CATS-HED} & RGB &\textbf{0.732} &\textbf{0.746} & \textbf{0.405} & \textbf{0.418}\\
			&HHA&\textbf{0.693}&\textbf{0.703} & \textbf{0.349} & \textbf{0.362}\\
			&RGB-HHA& \textbf{0.755} & \textbf{0.770} & \textbf{0.382} &\textbf{0.397}\\
			\hline
			\multirow{3}{*}{RCF\cite{liu2019richer}} & RGB &0.745 &0.759 &0.398 &0.413\\
			&HHA&0.701 &0.712 &0.333 &0.348\\
			&RGB-HHA& 0.764 & 0.778 & 0.374 & 0.385\\
			\hline
			\multirow{3}{*}{CATS-RCF} & RGB &\textbf{0.752} &\textbf{0.765} &\textbf{0.474} &\textbf{0.488} \\
			&HHA&\textbf{0.710}&\textbf{0.721} &\textbf{0.433} &\textbf{0.445}\\
			&RGB-HHA& \textbf{0.768} &\textbf{ 0.782} &\textbf{0.439} &\textbf{0.452}\\
			\hline
			\multirow{3}{*}{BDCN\cite{he2019bi-directional}} & RGB &0.748  &0.762 &0.426 &0.450\\
			&HHA&0.704&0.716 &0.347 &0.367\\
			&RGB-HHA& 0.766 & 0.779 &0.375 &0.392\\
			\hline
			\multirow{3}{*}{CATS-BDCN} & RGB &\textbf{0.752} &\textbf{0.765} &\textbf{0.442} &\textbf{0.462}\\
			&HHA & \textbf{0.712} & \textbf{0.724} & \textbf{0.422} & \textbf{0.439}\\
			&RGB-HHA& \textbf{0.770} & \textbf{0.783} & \textbf{0.418} & \textbf{0.435}\\
			\hline
		\end{tabular}
	\vspace{-1mm}
	\end{table}
	
\begin{figure*}[!t]
	\setlength{\fboxsep}{0pt}
	\begin{tabular}{cccccc}

		\fbox{\includegraphics[width=0.14\linewidth]{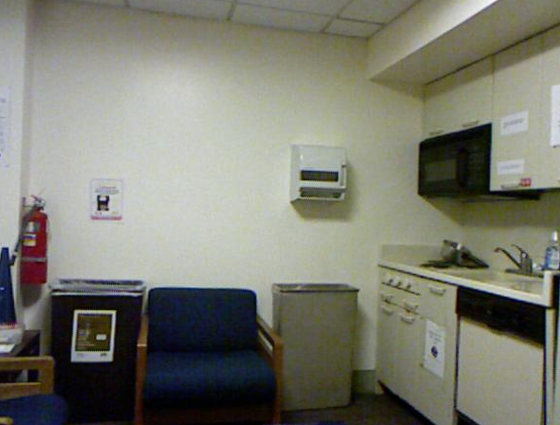}}&
		\fbox{\includegraphics[width=0.14\linewidth]{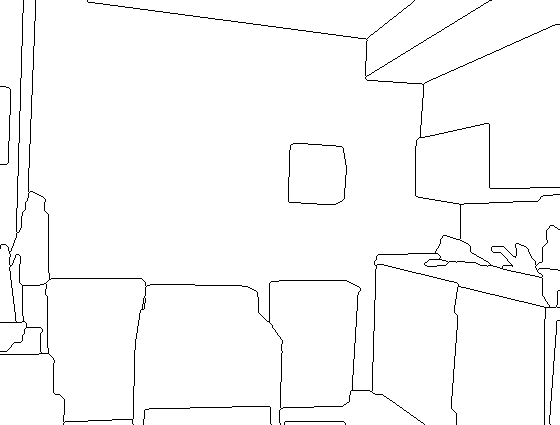}}&
		\fbox{\includegraphics[width=0.14\linewidth]{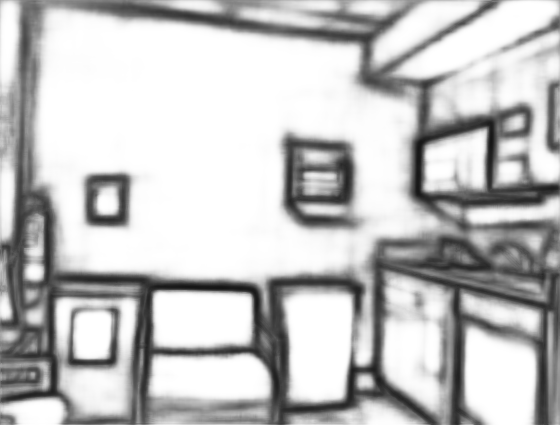}}&
		\fbox{\includegraphics[width=0.14\linewidth]{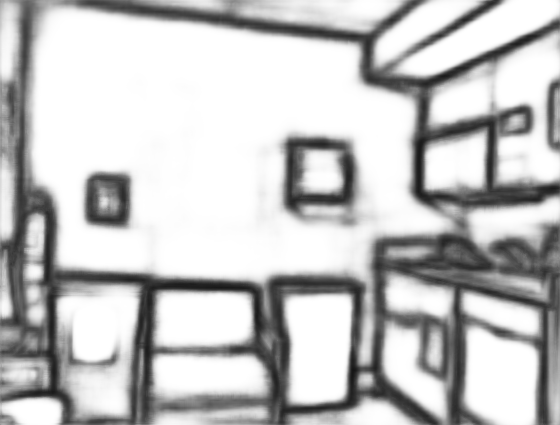}}&
		\fbox{\includegraphics[width=0.14\linewidth]{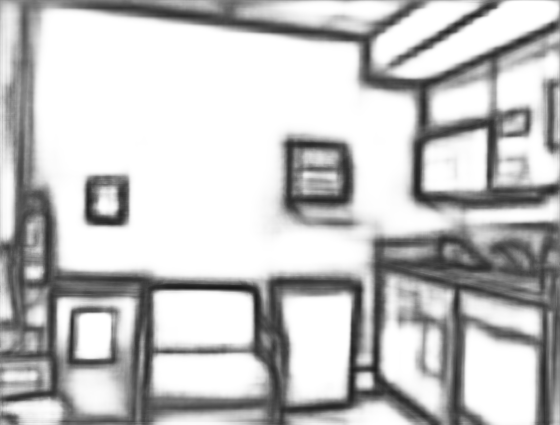}}&
		\fbox{\includegraphics[width=0.14\linewidth]{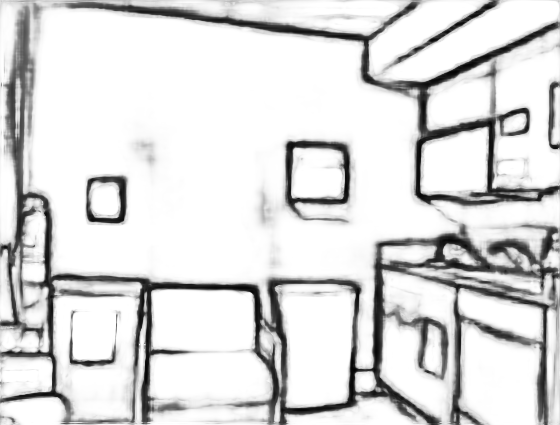}}
		
		\\

		\fbox{\includegraphics[width=0.14\linewidth]{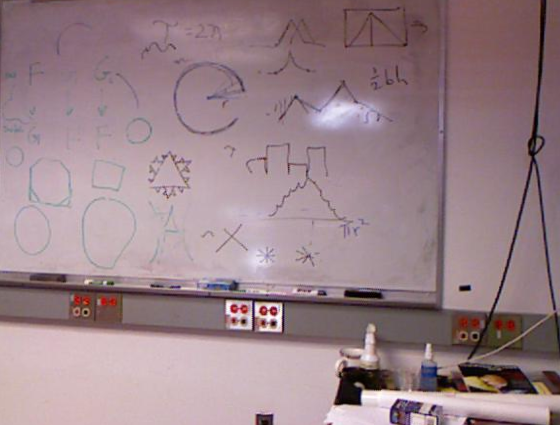}}&
		\fbox{\includegraphics[width=0.14\linewidth]{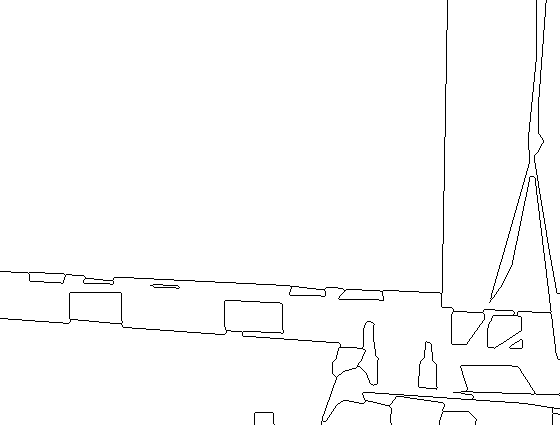}}&
		\fbox{\includegraphics[width=0.14\linewidth]{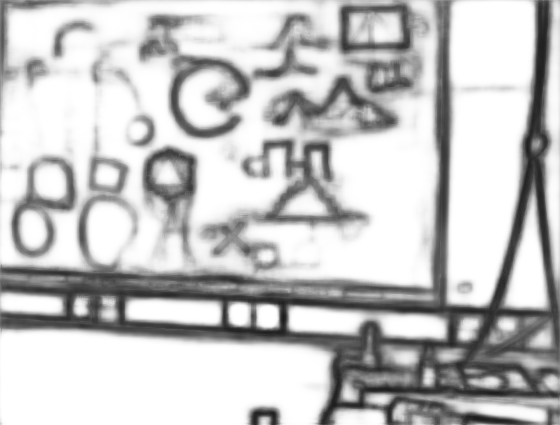}}&
		\fbox{\includegraphics[width=0.14\linewidth]{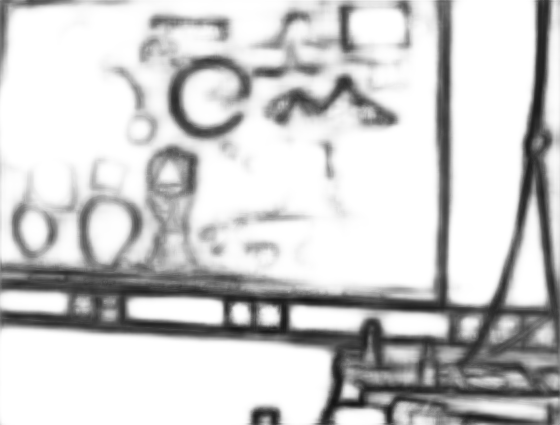}}&
		\fbox{\includegraphics[width=0.14\linewidth]{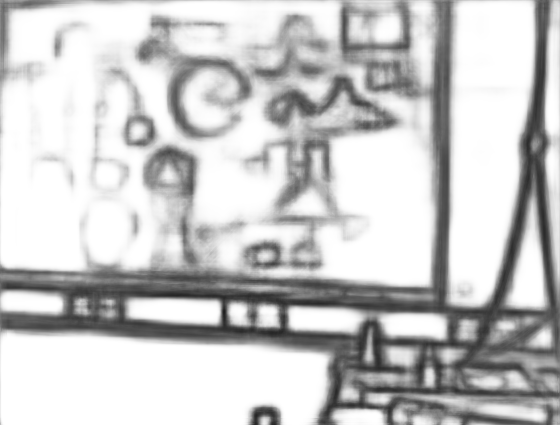}}&
		\fbox{\includegraphics[width=0.14\linewidth]{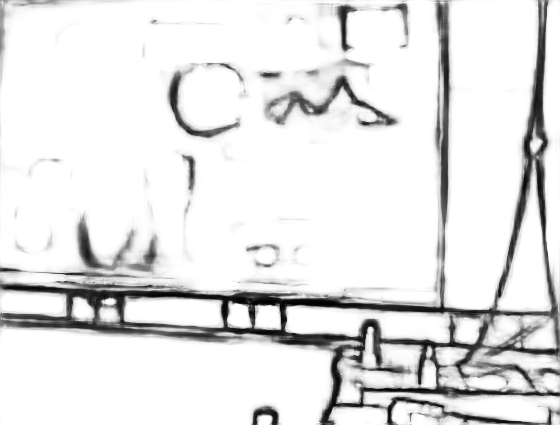}}
		
		\\

		\fbox{\includegraphics[width=0.14\linewidth]{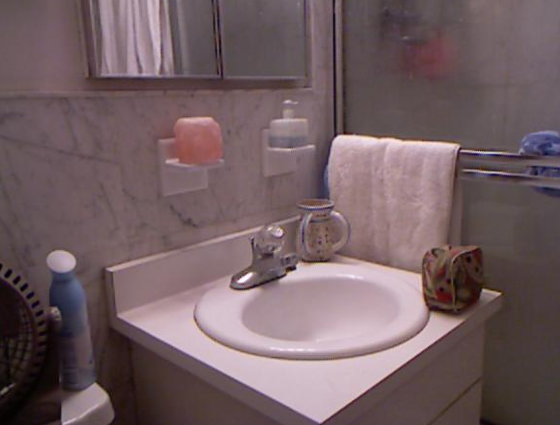}}&
		\fbox{\includegraphics[width=0.14\linewidth]{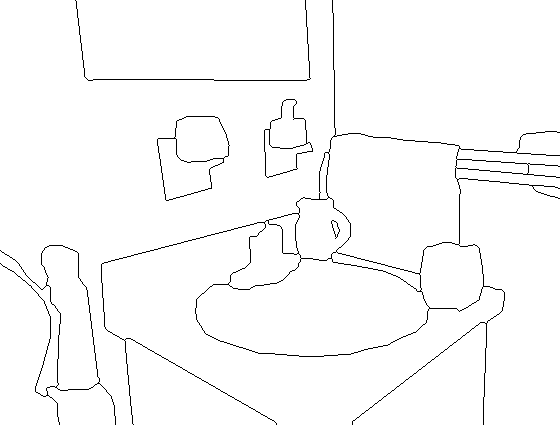}}&
		\fbox{\includegraphics[width=0.14\linewidth]{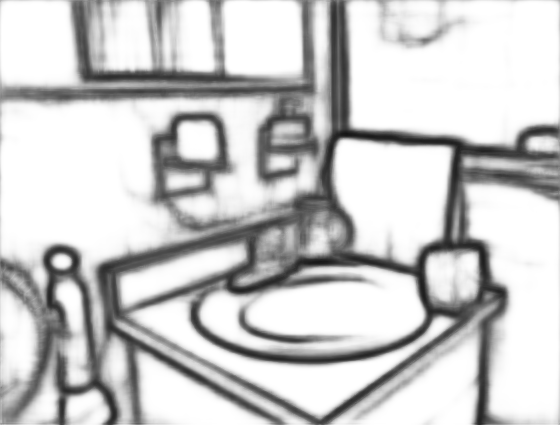}}&
		\fbox{\includegraphics[width=0.14\linewidth]{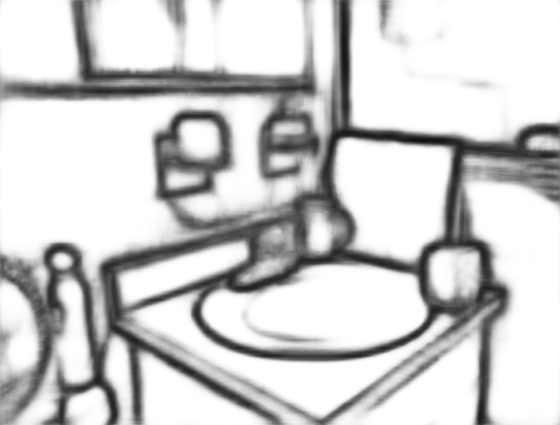}}&
		\fbox{\includegraphics[width=0.14\linewidth]{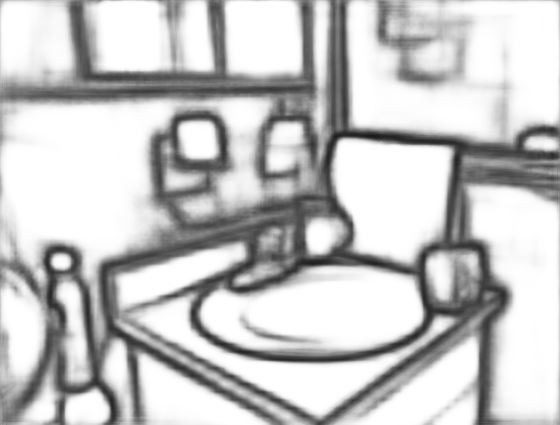}}&
		\fbox{\includegraphics[width=0.14\linewidth]{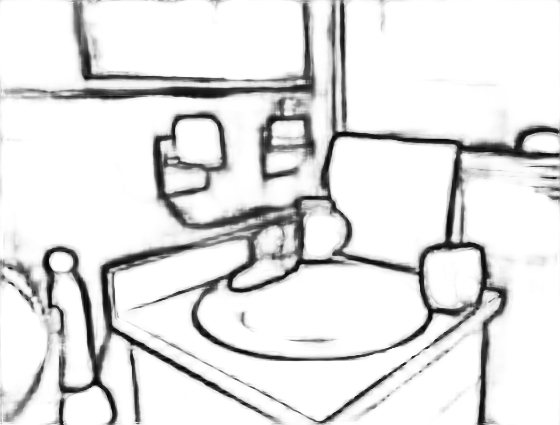}}
		
		\\

		\fbox{\includegraphics[width=0.14\linewidth]{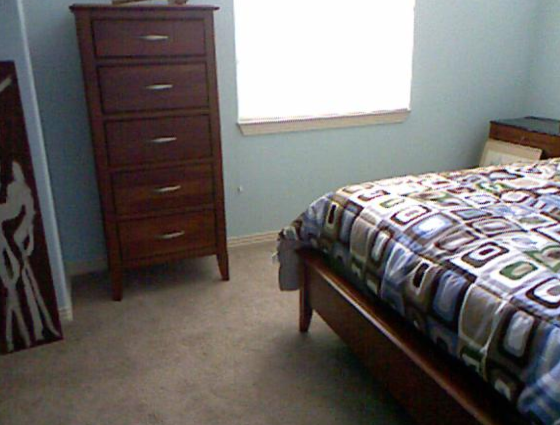}}&
		\fbox{\includegraphics[width=0.14\linewidth]{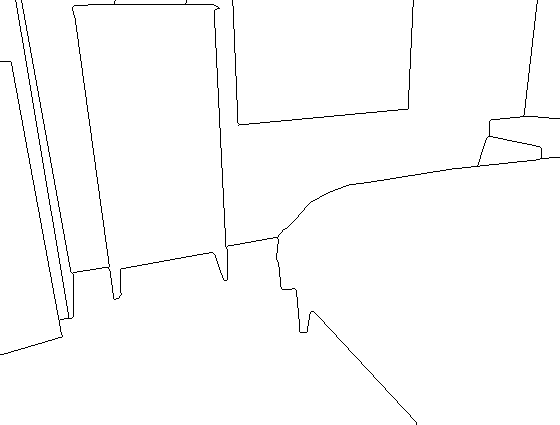}}&
		\fbox{\includegraphics[width=0.14\linewidth]{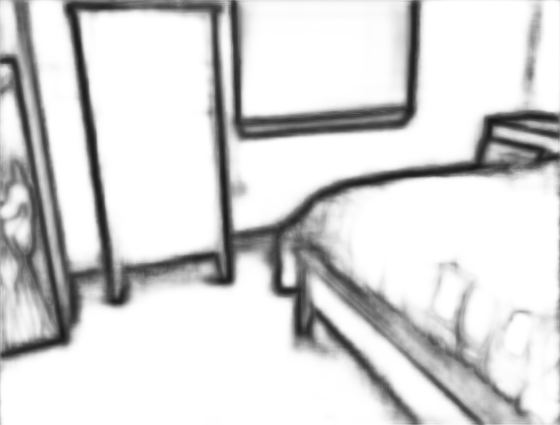}}&
		\fbox{\includegraphics[width=0.14\linewidth]{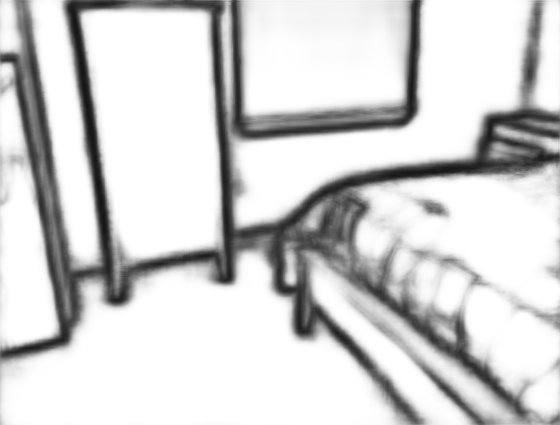}}&
		\fbox{\includegraphics[width=0.14\linewidth]{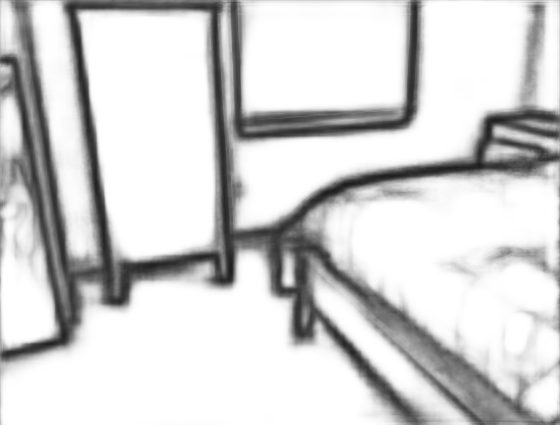}}&
		\fbox{\includegraphics[width=0.14\linewidth]{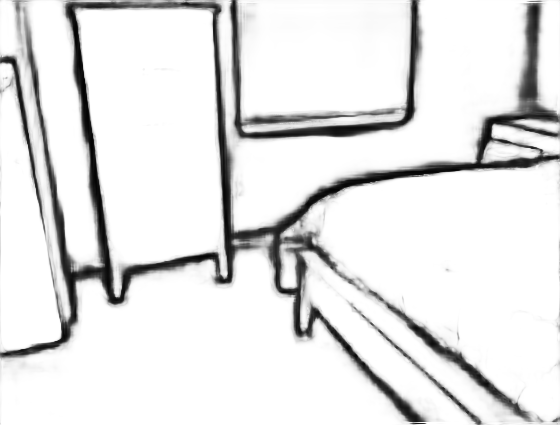}}
		
		\\
		
		Image & Label & HED~\cite{xie2017hed} & RCF~\cite{liu2019richer} & BDCN~\cite{he2019bi-directional} & CATS-RCF
	\end{tabular}
	\vspace{-2mm}
	\caption{Qualitative comparison between the prior arts of deep edge detection and our proposed CATS on the NYUDv2 dataset~\cite{Silberman:ECCV12}.}
	\vspace{-4mm}
	\label{fig:vis_nyud}
\end{figure*}

	Compared to the original HED, RCF and BDCN, their CATS versions all show steady accuracy growth regarding the standard evaluation, where the gain in ODS and OIS can respectively reach as much as 1.0\% and 0.9\% with respect to the RGB data. The crispness-emphasized evaluation also reveals the remarkable improvement in edge localization accuracy by the proposed CATS, which brings 7.6\% and 7.5\% progress at most in the ODS and OIS scores respectively for the RGB data. In \figref{fig:vis_nyud}, we present the qualitative results on the NYUDv2 dataset, where the proposed CATS generates crisper edges than other models, illustrating the capability of CATS for robust texture suppression in complex circumstances.
	For a comprehensive comparison, \figref{fig:pr} plots the precision-recall curves of the CATS-based models and the original networks.	
	
	\subsubsection{Multicue Dataset}
	The CATS-applied models are further compared with HED, RCF and BDCN on the edge and boundary data of Multicue via the standard evaluation procedure. In evaluation, the performance of a model is measured by the average score of three independent trials, and the performance fluctuation is appraised by the mean standard deviation. The quantitative results with only single-scale input are presented in \tabref{tab:multicue_study}.
	\begin{table}[!h]
		\centering
		\vspace{-3mm}
		\caption{Evaluation results on Multicue dataset.The mean standard deviations are given in the parentheses.}
		\label{tab:multicue_study}
		\vspace{-3mm}
		\begin{tabular}{c|c||c|c}
			\hline
			Category & Method & ODS & OIS\\
			\hline
			\multirow{8}{*}{Boundary} & Human\cite{M2016A} & 0.760(0.017) & - \\
			& Multicue\cite{M2016A} & 0.720(0.014) & -\\
			& HED\cite{xie2017hed} & 0.814(0.011) & 0.822(0.008)\\
			& CATS-HED & \textbf{0.827(0.007)} & \textbf{0.830(0.008)}\\
			& RCF\cite{liu2019richer} & 0.817(0.004) & 0.825(0.005)\\
			& CATS-RCF & \textbf{0.841(0.001)} & \textbf{0.846(0.002)}\\
			& BDCN\cite{he2019bi-directional} & 0.836(0.001) & 0.846(0.003)\\
			& CATS-BDCN & \textbf{0.842(0.001)} & \textbf{0.847(0.001)}\\
			\hline
			\multirow{8}{*}{Edge} & Human\cite{M2016A} & 0.750(0.024) & -\\
			& Multicue\cite{M2016A} & 0.830(0.002) & -\\
			& HED\cite{xie2017hed} & 0.851(0.014) & 0.864(0.011)\\
			& CATS-HED & \textbf{0.885(0.004)} & \textbf{0.893(0.003)}\\
			& RCF\cite{liu2019richer} & 0.857(0.004) & 0.862(0.004)\\
			& CATS-RCF & \textbf{0.892(0.001)} & \textbf{0.895(0.001)}\\
			& BDCN\cite{he2019bi-directional} & 0.891(0.001) & 0.898(0.002)\\
			& CATS-BDCN & \textbf{0.897(0.001)} & \textbf{0.904(0.001)}\\
			\hline
		\end{tabular}
		\vspace{-1mm}
	\end{table}
	
	In \tabref{tab:multicue_study}, the CATS-applied approaches obtain an apparent detection accuracy increase against the corresponding original models on both the Multicue Boundary and Multicue Edge datasets.
	
	\subsection{Ablation Study}
	The ablation study was conducted on NYUDv2 dataset to avoid the impact of the label threshold, and the NYUDv2 dataset was only augmented by flipping. RCF~\cite{liu2019richer} with weighted cross entropy was chosen as the baseline for comparison. We verified the functionality of each component in the tracing loss, and investigated the side unmixing effect of the CoFusion block by substituting it for the weighted average layer in RCF. All models were conventionally trained from scratch, where the weights in convolution layers were initialized by zero-mean Gaussian distributions with a standard deviation 0.01 and the biases were set to 0. \tabref{tab:ablation} presents the details of quantitative analysis.
	
	\begin{table}[!ht]
		\vspace{-3mm}
		\centering
		\caption{The numerical results for ablation study. $L_{bdry}$ and $L_{tex}$ denote the boundary tracing function and texture suppression function. SEval and CEval respectively denote the standard evaluation and the crispness-emphasized evaluation.}
		\label{tab:ablation}
		\vspace{-0.1in}
		\scriptsize
		\begin{tabular}{cccc|p{4mm}<{\centering}c|p{4mm}<{\centering}c}
			\hline
			\multirow{2}{*}{} & \multicolumn{3}{c}{CATS} & \multicolumn{2}{c}{SEval} & \multicolumn{2}{c}{CEval} \\ \cline{2-8}
			& $L_{bdry}$ & $L_{tex}$ & CoFusion & ODS & OIS & ODS & OIS  \\
			\hline
			RCF & & & & 0.720 & 0.736 & 0.381 & 0.397 \\
			RCF & $\checkmark$ & & & 0.722 & 0.738 & 0.382 & 0.399 \\
			RCF & & $\checkmark$ & & 0.726 & 0.743 & 0.383 & 0.396  \\
			RCF & $\checkmark$ & $\checkmark$ & & 0.729 & 0.744 & 0.411 & 0.424 \\
			RCF & & & $\checkmark$ & 0.725 & 0.739 & 0.381 & 0.394  \\
			RCF & $\checkmark$ & & $\checkmark$ & 0.727 & 0.740 & 0.427 & 0.441 \\
		    RCF & & $\checkmark$ & $\checkmark$ & 0.728 & 0.743 & 0.409 & 0.427\\
			CATS-RCF & $\checkmark$ & $\checkmark$ & $\checkmark$ & 0.731 & 0.744 & 0.451 & 0.468\\
			\hline
		\end{tabular}
	\end{table}

	In \tabref{tab:ablation}, the complete CATS improves the performance of RCF by 1.1\% in ODS and 0.8\% in OIS, while using the tracing loss or the CoFusion block alone gains over the original RCF by at most 0.9\% in ODS and 0.8\% in OIS.

	In terms of the crispness-emphasized evaluation displayed in \tabref{tab:ablation}, the proposed CATS gains over the RCF by 7\% in ODS and 7.1\% in OIS. The results in the first three rows show that, although $L_{bdry}$ or $L_{tex}$ alone can promote the performance of RCF with post-processing, they still have similar crispness scores with the weighted cross entropy. This is because that $L_{bdry}$ and $L_{tex}$ focus on different types of non-edge points. Only applying one of them is insufficient for a comprehensive refinement of side edges. In contrast, with the effective feature unmixing and texture suppression offered by combining $L_{bdry}$ and $L_{tex}$, the tracing loss brings a remarkable crispness enhancement to RCF. Similarly, the CoFusion block guided by the weighted cross entropy also increases the scores of the standard evaluation measures, but with little crispness improvement. When working with $L_{bdry}$ or $L_{tex}$, however, the CoFusion block shows its power in addressing side mixing for crisp edge detection. With the complete CATS where the CoFusion is guided by the tracing loss, CATS-RCF achieves the most crispness improvement, which demonstrates that the tracing loss and the CoFusion block work complementarily for crisp edge detection.

	\begin{figure*}[t]
	\centering
	\begin{tabular}{cccccc}
		\includegraphics[width=0.16\linewidth]{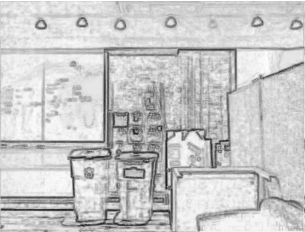}
		& \includegraphics[width=0.16\linewidth]{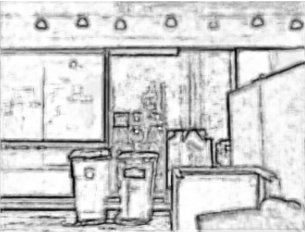}
		& \includegraphics[width=0.16\linewidth]{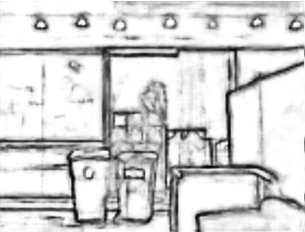}
		& \includegraphics[width=0.16\linewidth]{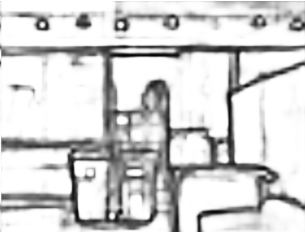}
		& \includegraphics[width=0.16\linewidth]{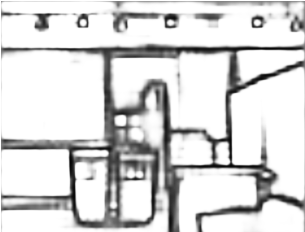} &  \multirow{2}{0.2in}[0.755in]{\includegraphics[width=0.28in]{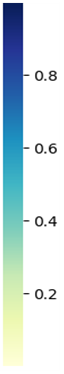}}
		
		\\

		\includegraphics[width=0.16\linewidth]{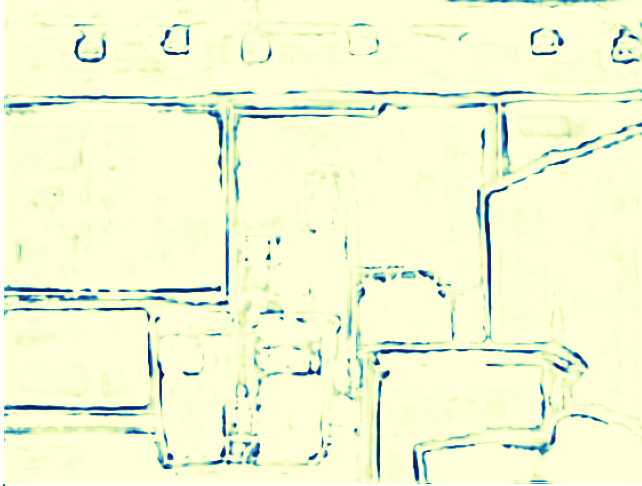}
		& \includegraphics[width=0.16\linewidth]{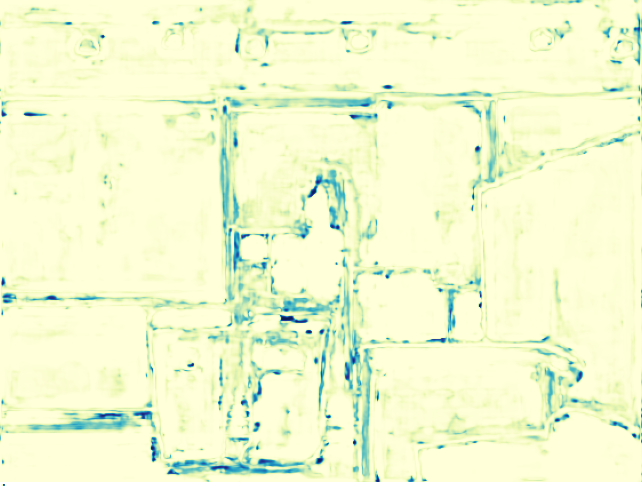}
		& \includegraphics[width=0.16\linewidth]{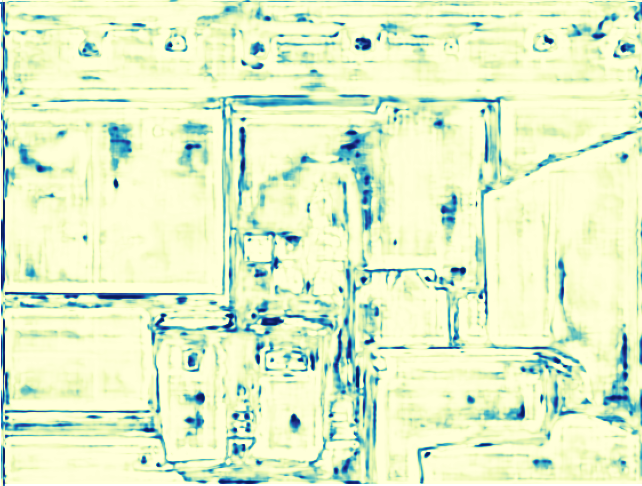}
		& \includegraphics[width=0.16\linewidth]{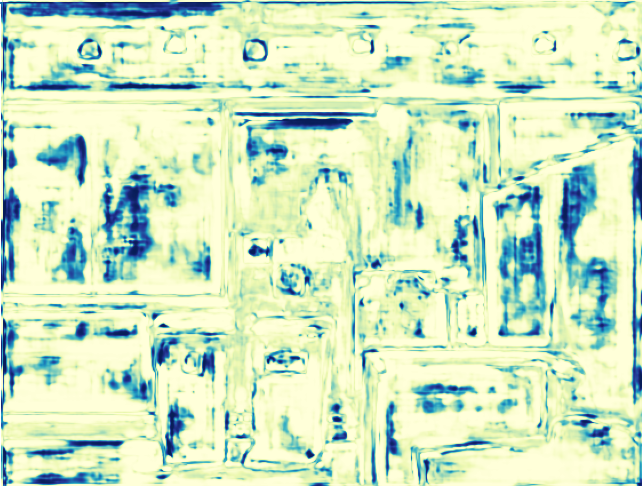}
		& \includegraphics[width=0.16\linewidth]{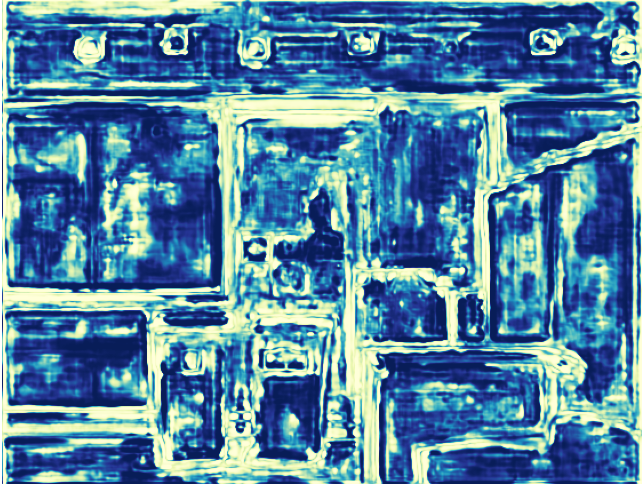} &
		
		\\
		
		Side Edge$_1$ & Side Edge$_2$ & Side Edge$_3$ & Side Edge$_4$ & Side Edge$_5$ & 
	\end{tabular}
	
	\vspace{-1mm}
	\caption{The weight maps generated by the CoFusion block in CATS-RCF for multi-level side edges on the NYUDv2 dataset~\cite{Silberman:ECCV12}. The results show the different attention preference of multi-level side edges for non-edge region smoothing and edge details.}
	\vspace{-4mm}
	\label{fig:attnvis}
\end{figure*}

	\subsection{Deeper into the CoFusion block}

	For a perceptual understanding of the side unmixing effect in the CoFusion block, we visualized the weight maps generated by the CoFusion block in \figref{fig:attnvis}.

	It can be found that the CoFusion block gives more attention to the two highest-level outputs on the finely smoothed non-edge regions and higher weights to the lower-level side predictions at boundaries. By concentrating on the different merits of side maps for edge delineation, the CoFusion block solves the issue of side mixing in the fusion procedure. 
	
	\vspace{-4mm}
	\section{Conclusion}
	\label{sec:conclusion}
    In this paper, we introduce a simple yet effective context-aware tracing strategy (CATS) to tackle the problem of localization ambiguity for modern deep edge detectors. The CATS addresses this issue by unmixing the edge features with a tracing loss and the side predictions with a CoFusion block during the fusion procedure. The effectiveness of CATS was validated in three VGG16-based edge detectors, \ie{HED, RCF, and BDCN}. Extensive experiments demonstrate that the CATS can bring consistent performance improvement to the three detectors. Especially, the qualitative and the crispness-emphasized evaluation results show that the CATS facilitates modern edge detectors to obtain crisp edges from raw edge predictions, with the localization accuracy dramatically improved. 
    In the future, it would be interesting to explore the effectiveness of CATS for other down-stream vision tasks. Source code and more results are available at \url{https://github.com/WHUHLX/CATS}.
	
	\vspace{-4pt}
	\section*{Acknowledgment}
	This research was funded by the National Key Research and Development Program of China under Grant 2018YFB0505401, and the National Natural Science Foundation of China Project under Grant 42071370.
	
	
	%


	
	
	{
	\vspace{-4mm}
	\small 
	\bibliographystyle{IEEEtran}
	\bibliography{ref}
	}
	%
	
	
	
	%
	
\end{document}